\documentclass{article} 
\usepackage{morse_preprint,times}

\usepackage{amsmath,amsfonts,bm}

\def\eqref#1{equation~\ref{#1}}

\def\1{\bm{1}}

\DeclareMathAlphabet{\mathsfit}{\encodingdefault}{\sfdefault}{m}{sl}
\SetMathAlphabet{\mathsfit}{bold}{\encodingdefault}{\sfdefault}{bx}{n}

\usepackage{bbding}
\usepackage{pifont}

\usepackage[utf8]{inputenc} 
\usepackage{wasysym} 
\DeclareUnicodeCharacter{266B}{\twonotes}
\DeclareUnicodeCharacter{2026}{\ldots}
\usepackage[T1]{fontenc}    
\usepackage{hyperref}       
\usepackage{url}            
\hypersetup{pdftitle={MORSE: Multi-Context Ordering via Reverse Scoring for Evidence-Preserving Compression}, pdfauthor={Ke Wan, Yifan Wang, Liheng Lai, Chen Chen}}
\usepackage{amsfonts}       
\usepackage{nicefrac}       
\usepackage{algorithmicx}   
\usepackage{microtype}
\usepackage{needspace} 
\PassOptionsToPackage{table,xcdraw}{xcolor}
\usepackage{xcolor}         
\usepackage{siunitx}
\definecolor{lightcrimson}{rgb}{0.93, 0.16, 0.51}

\usepackage{amsmath, amssymb, amsthm}

\usepackage{booktabs}
\usepackage{threeparttable}
\usepackage{graphicx, subcaption, adjustbox, wrapfig, multirow, array, tabularx, enumitem, ulem,colortbl}
\usepackage{xcolor, hyperref}
\usepackage{tikz}
\usepackage{algorithm}
\usepackage{algpseudocode}
\usepackage{amsthm}


\theoremstyle{plain}

\title{MORSE: Multi-Context Ordering via Reverse Scoring for Evidence-Preserving Compression}

\newcommand{\MORSEAuthorGrid}{%
  \parbox[t]{\dimexpr\textwidth-2\tabcolsep\relax}{%
    \noindent
    \begin{minipage}[t]{0.45\linewidth}
      \raggedright\normalfont
      \textbf{Ke Wan}\\
      Department of Computer Science\\
      University of Virginia\\
      Charlottesville, VA, USA\\
      \texttt{tbn5pj@virginia.edu}
    \end{minipage}%
    \hfill
    \begin{minipage}[t]{0.45\linewidth}
      \raggedright\normalfont
      \textbf{Yifan Wang}\\
      AfterQuery\\
      \texttt{yifan@afterquery.com}
    \end{minipage}%
    \par\vspace{1.25em}
    \noindent
    \begin{minipage}[t]{0.45\linewidth}
      \raggedright\normalfont
      \textbf{Liheng Lai}\\
      AfterQuery\\
      \texttt{liheng@afterquery.com}
    \end{minipage}%
    \hfill
    \begin{minipage}[t]{0.45\linewidth}
      \raggedright\normalfont
      \textbf{Chen Chen\thanks{Corresponding author.}}\\
      Department of Computer Science\\
      University of Virginia\\
      Charlottesville, VA, USA\\
      \texttt{zrh6du@virginia.edu}
    \end{minipage}%
  }%
}
\author{\MORSEAuthorGrid}

\begin{document}
\maketitle

\begin{abstract}


Retrieval-augmented generation often relies on multiple retrieved contexts that contain substantial redundancy, motivating context compression to preserve useful information under limited input budgets. Likelihood-based compressors can account for cross-context redundancy through sequential scoring, but this makes evidence scores dependent on context order. We show that permuting the same contexts under an unchanged compressor can substantially change which supporting evidence survives compression. We attribute this sensitivity to information preemption: earlier, partially relevant contexts can absorb credit for shared information, reducing the incremental scores of later, stronger evidence and increasing its risk of removal.
Controlled pair-swap interventions provide direct empirical support for this mechanism by showing that placing stronger evidence before overlapping, partially relevant contexts can improve its survival.
Based on this insight, we introduce \texttt{MORSE}, a compression-aware method for evidence-preserving context ordering. \texttt{MORSE} uses reverse query likelihood to construct an evidence-first anchor and to evaluate compressed candidate outputs, enabling compression-aware selection among alternative permutations.
Across multi-hop Question Answering (QA) benchmarks, compression procedures, budgets, and scoring models,
\texttt{MORSE} improves evidence retention over reverse
ordering and generally outperforms matched random
search, with downstream QA gains.
Our code is available at \url{https://github.com/tbn5pj/MORSE_code}.

\end{abstract}

\section{Introduction}

Large language models increasingly rely on collections of retrieved contexts for knowledge-intensive
question answering (QA)~\citep{arslan2024survey,NEURIPS2020_6b493230}.
While retrieval improves information coverage, the resulting inputs often contain substantial redundancy, partial overlap, and weakly relevant passages.
Context compression addresses this inefficiency by retaining only the text units most useful for answering the query under a limited input budget \citep{xu-etal-2024-concise,li2023compressing,jiang2023llmlingua,li2025prompt}.
A prominent family of extractive compression methods uses 
language-model likelihoods to estimate unit importance under a target budget, with later methods incorporating query-aware signals to better preserve task-relevant information~\citep{jiang2023llmlingua,jiang2024longllmlingua,li2023compressing,cao2024retaining}.
Intuitively, query-conditioned likelihood scoring measures how much a query increases the predictability of a text unit beyond what the preceding text already explains.

In this work, we focus on \emph{query-conditioned} compression, where the downstream query is available when estimating what to preserve.
In multi-context inputs, 
naive approaches that independently estimate each context's relevance cannot account for redundancy across contexts \citep{hwang2025exit,jeong2025ecorag}: overlapping passages may each receive credit for the same information, causing a limited compression budget to be spent repeatedly on redundant evidence.
Sequential conditioning provides a natural way to account for such overlap by evaluating new information relative to what has already appeared.
Yet this introduces a new degree of freedom: because the preceding information depends on context order, the resulting importance scores become order-dependent.


\begin{figure*}[t]
\centering
\begin{minipage}[c]{0.35\textwidth}    
\centering    
\label{fig:ordering_sensitivity}    
\includegraphics[width=\linewidth]{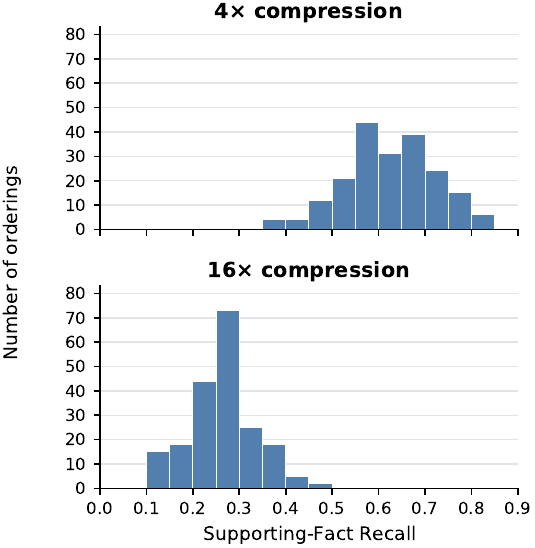}
    \vspace{0.5mm}    
\textbf{(a) Ordering sensitivity}
\end{minipage}
\hfill
\begin{minipage}[c]{0.61\textwidth}    
\centering    
\includegraphics[width=\linewidth]{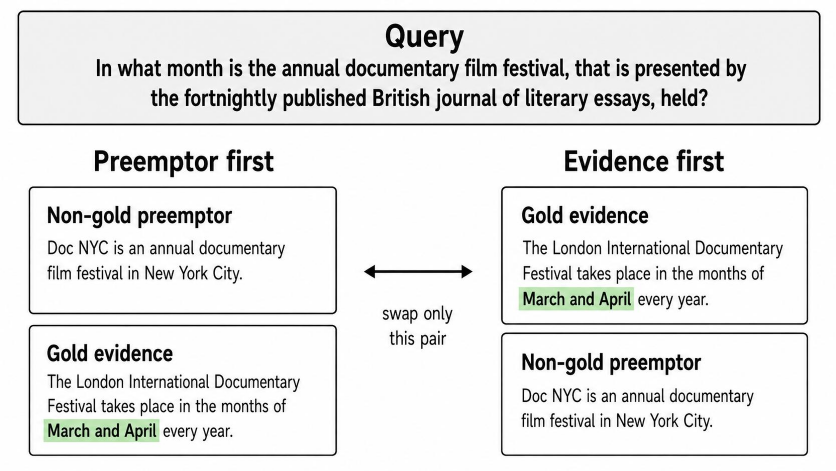}
    \vspace{0.5mm}    
\textbf{(b) Information preemption}
\end{minipage}
\caption{
\textbf{Context ordering strongly affects evidence retention and can induce information preemption.}
\textbf{(a)} On HotpotQA, we evaluate 200 random permutations of the same retrieved context sets.
Under both $4\times$ and $16\times$ compression, Supporting-Fact Recall (SF-R) varies substantially across orderings despite using identical contexts and the same compressor, demonstrating that context order is a consequential decision variable in sequential compression.
\textbf{(b)} A concrete pair-swap example illustrates one mechanism behind this sensitivity.
The non-gold preemptor overlaps with the gold evidence but does not contain the answer-bearing fact, \emph{March and April}.
When the preemptor appears first, the gold supporting sentence receives score $0.45$, ranks $32$nd, and is deleted at all three compression ratios.
Swapping only this pair raises its score to $9.04$, moves it to rank $2$, and preserves it at $4\times$, $8\times$, and $16\times$ compression, increasing SF-R from $0.5$ to $1.0$.
}
\label{fig:ordering_motivation}
\end{figure*}

We find that this ordering dependence can be substantial.
As shown in Figure~\ref{fig:ordering_motivation}, different permutations of exactly the same retrieved contexts lead to markedly different evidence-retention outcomes, even when the query, scoring model, compression budget, and deletion procedure are all held fixed.
Unlike prior work that reorders contexts to improve downstream language-model performance~\citep{jiang2024longllmlingua}, we study how context ordering affects which evidence survives compression itself.
This reveals context ordering as a consequential decision variable in likelihood-based multi-context compression, with considerable headroom beyond the original retrieval order and random ordering.
We therefore treat evidence retention as the primary outcome of interest: unlike downstream answer accuracy, it directly isolates ordering's effect on the compression process itself.

We attribute this sensitivity to an order-dependent credit-assignment mechanism in sequential likelihood scoring.
When a text unit is evaluated, its score depends on the information already present in the preceding contexts.
Earlier information can make overlapping content more predictable, reducing the incremental score assigned to later contexts.
This creates a failure mode that we call \emph{information preemption}: a partially relevant context may appear first and absorb credit for information shared with a later, more complete evidence carrier.
The stronger evidence carrier can then be undervalued and disproportionately removed under a limited compression budget.

Motivated by this observation, we introduce \texttt{MORSE}:
\textbf{M}ulti-Context \textbf{O}rdering via
\textbf{R}everse \textbf{S}coring for
\textbf{E}vidence-Preserving Compression.
\texttt{MORSE} uses reverse query scoring, which measures how much a text increases the language model's likelihood of the query, to guide context ordering at two levels.
First, it independently scores each context and constructs an evidence-first anchor by placing contexts with stronger standalone query evidence earlier.
However, standalone scores cannot fully capture the redundancy, complementarity, and preemption induced by sequential compression.
\texttt{MORSE} therefore evaluates the anchor alongside alternative global permutations using the actual compressor and target budget.
It then applies the same reverse scoring principle to the resulting compressed outputs and selects the candidate that retains the strongest estimated evidence for the query.

\paragraph{Contributions.}
We summarize our contributions as follows:
(i) We uncover substantial ordering sensitivity in likelihood-based multi-context compression, showing that different permutations of the same retrieved contexts can lead to markedly different evidence-retention outcomes.
(ii) We identify \emph{information preemption} as an order-dependent credit-assignment failure mode and provide direct empirical support through controlled pair-swap interventions that alter only the relative order of measured preemptors and evidence carriers.
(iii) We introduce \texttt{MORSE}, a compression-aware context-ordering method that applies a common reverse query-evidence principle to both standalone contexts and compressed outputs, combining an evidence-first anchor with global permutation exploration and compression-aware selection.

\section{Related Work}

\textbf{Likelihood-Based Context Compression.}
Likelihood-based context compression uses language-model probabilities to identify and remove less informative content. Selective Context employs self-information to prune redundant lexical units, while \textbf{LLMLingua} develops a coarse-to-fine framework with iterative likelihood-based scoring to capture dependencies among compressed content \citep{li2023compressing,jiang2023llmlingua}. \textbf{LongLLMLingua} further introduces question-aware compression and document reordering for long-context inputs, and subsequent work explores query-guided or learned compression objectives to better preserve task-relevant information \citep{jiang2024longllmlingua,cao2024retaining,pan2024llmlingua,xu2024recomp,ge2023context,li2025500xcompressor,shandilya2025taco,nagle2024fundamental,xu-etal-2024-concise}. In contrast to designing a new unit-level scoring or deletion rule, our work studies context ordering as a structural variable in sequential likelihood-based compression, showing that different permutations alter how shared evidence is credited and, consequently, which information survives compression.

\textbf{Context Ordering for Long-Context Language Models.}
Context order is known to affect how effectively language models use long inputs \citep{hsieh2024found,zhang2024found,tang2024found}. \citet{liu2024lost} show that long-context LMs exhibit strong positional sensitivity, with relevant information often being used less effectively when placed in the middle of the input. Motivated by this phenomenon, \textbf{LongLLMLingua} reorders retrieved documents to place important information at favorable positions, and subsequent work similarly explores retrieval reordering or positional-bias mitigation to improve downstream generation in long-context RAG \citep{jiang2024longllmlingua,jin2025long,cuconasu-etal-2025-rag}. These studies primarily treat ordering as a way to control how information is presented to the downstream language model. In contrast, we study ordering as a variable within the compression process itself: context permutation changes sequential likelihood histories, altering evidence credit assignment and ultimately determining which information survives compression.


\textbf{Redundancy-Aware Retrieval and Evidence Selection.}
Redundancy has long been studied in information retrieval and evidence selection.
Diversification methods such as \textbf{xQuAD} seek to cover complementary aspects of an information need \citep{santos2010explicit}.
Similar ideas have recently been applied to retrieval-augmented generation, where diversity-aware retrieval selects complementary evidence to reduce redundancy in the retrieved context set \citep{khan2026df}.
These approaches primarily address \emph{which} contexts should be retrieved or retained. In contrast, \texttt{MORSE} keeps the retrieved context set fixed and studies \emph{how} those contexts should be ordered before sequential compression, where redundancy induces order-dependent evidence attribution even when the selected information itself is unchanged.
\section{Problem Formulation}
\label{sec:problem}

Let $q$ denote a query and $\mathcal{C}=\{c_1,\ldots,c_n\}$ a collection of retrieved contexts, where each context $c_i=(u_{i,1},\ldots,u_{i,m_i})$ consists of ordered compression units.
We keep the formulation unit-agnostic; in our experiments, each compression unit is a complete sentence.
Let $L_\theta(u\mid h)=-\log p_\theta(u\mid h)$ denote the code length of unit $u$ conditioned on history $h$.
We consider the query-conditioned importance score
\begin{equation}
    s(u\mid q,h)
    =
    L_\theta(u\mid h)
    -
    L_\theta(u\mid q,h).
    \label{eq:qmi_score}
\end{equation}

For sequential compression, let $\pi=(\pi_1,\ldots,\pi_n)$ denote a permutation of the retrieved contexts, giving $\mathcal{C}_\pi=(c_{\pi_1},\ldots,c_{\pi_n})$.
For unit $u_{\pi_t,j}$, its preceding history is
$H_\pi(u_{\pi_t,j})=(c_{\pi_1},\ldots,c_{\pi_{t-1}},u_{\pi_t,<j})$,
and its sequential importance score is
\begin{equation}
    s_\pi(u_{\pi_t,j})
    =
    s\!\left(
        u_{\pi_t,j}
        \mid
        q,H_\pi(u_{\pi_t,j})
    \right).
    \label{eq:sequential_score}
\end{equation}
We refer to $s_\pi$ as the sequential chain query-mutual-information (QMI) score.
Under token budget $B$, the compressor removes low-scoring units according to its deletion procedure and produces an extractive output
$\widetilde{\mathcal{C}}_{\pi,B}$
satisfying
$\sum_i |\widetilde{c}_{\pi_i}|\leq B$.
Because $s_\pi$ depends on the permutation-dependent history $H_\pi$, different permutations can produce different compressed outputs even when the query, retrieved contexts, compressor, and budget are fixed.
We therefore treat the context permutation $\pi$ as the decision variable in sequential multi-context compression.



\section{Analysis of Sequential Multi-Context Compression}
\label{sec:analysis}

Section~\ref{sec:problem} shows that context ordering changes the histories used in sequential likelihood scoring and can therefore change the resulting deletion decisions.
We next analyze one mechanism that produces this dependence and use it to motivate an evidence-first principle for context ordering.

\subsection{Order-Dependent Credit Assignment}
\label{sec:credit_assignment}

Recall that
$s(u\mid q,h)=L_\theta(u\mid h)-L_\theta(u\mid q,h)$.
Using $L_\theta(u\mid h)=-\log p_\theta(u\mid h)$ gives
\begin{equation}
    s(u\mid q,h)
    =
    \log
    \frac{p_\theta(u\mid q,h)}
         {p_\theta(u\mid h)}.
    \label{eq:pointwise_information}
\end{equation}
Thus, the score measures the query-dependent contribution of $u$ that remains after conditioning on history $h$.
This conditioning naturally discounts redundancy.
If a later unit $u_2$ provides no additional query-dependent information beyond an earlier unit $u_1$, such that
$p_\theta(u_2\mid q,u_1)=p_\theta(u_2\mid u_1)$,
then Equation~\ref{eq:pointwise_information} gives
$s(u_2\mid q,u_1)=0$.
Unlike independent scoring, sequential scoring can therefore avoid repeatedly assigning full credit to overlapping information.

The same dependence on history makes credit assignment order-dependent.
Using the same score with an entire context treated as the evaluated unit, define the directional suppression from $c_i$ to $c_j$ as
\begin{equation}
    \Delta_{i\rightarrow j}(q)
    =
    s(c_j\mid q)
    -
    s(c_j\mid q,c_i).
    \label{eq:credit_reduction}
\end{equation}
A positive $\Delta_{i\rightarrow j}(q)$ means that placing $c_i$ first reduces the query-dependent credit assigned to $c_j$ relative to evaluating $c_j$ independently.
Consider an idealized case where a partial context $c_p$ contains shared evidence $S$, while a stronger evidence carrier $c_e$ contains both $S$ and complementary evidence $T$.
If $c_p$ appears first and makes $S$ predictable, the later $c_e$ may be evaluated primarily through its residual contribution:
\begin{equation}
    s(c_e\mid q,c_p)
    \approx
    s(T\mid q,S)
    \approx
    s(c_e\mid q)-s(S\mid q).
    \label{eq:preemption}
\end{equation}
This idealized decomposition illustrates a failure mode we call \emph{information preemption}: a stronger evidence carrier can lose credit for information already expressed by an earlier, only partially relevant context.
Equation~\ref{eq:preemption} is a mechanism illustration rather than a claim that every ordering effect follows this decomposition; whether suppression changes retention depends on whether it is large enough to alter the budget-constrained deletion decision.
Section~\ref{sec:pair_swap} tests this mechanism directly by changing only the relative order of measured preemptor--evidence pairs.
The analysis suggests an \emph{evidence-first principle}: stronger evidence carriers should preferentially appear earlier in sequential compression.

\subsection{Evidence-First Ordering and Its Limits}
\label{sec:evidence_first}

To formalize this principle, for any text $x$ we define its model-based reverse query evidence as
\begin{equation}
    E_\theta(x;q)
    =
    L_\theta(q)
    -
    L_\theta(q\mid x)
    =
    \log p_\theta(q\mid x)-\log p_\theta(q).
    \label{eq:reverse_query_evidence}
\end{equation}
For a fixed query, $L_\theta(q)$ is constant, so ranking texts by $E_\theta(x;q)$ is exactly equivalent to ranking them by reverse query likelihood $\log p_\theta(q\mid x)$.
Under a coherent joint-probability interpretation, Equation~\ref{eq:reverse_query_evidence} corresponds to the pointwise mutual information between $x$ and $q$; operationally, we use it as a PMI-style measure of how strongly $x$ supports the query.
At the context level, setting $x=c_i$ yields the \emph{standalone query evidence} of $c_i$, since each context is evaluated independently of the other retrieved contexts.
A forward score instead evaluates $\log p_\theta(c_i\mid q)$.
Under the same joint-probability interpretation,
\begin{equation}
    \log p_\theta(c_i\mid q)
    =
    E_\theta(c_i;q)
    +
    \log p_\theta(c_i),
    \label{eq:forward_decomposition}
\end{equation}
so forward likelihood also reflects the intrinsic likelihood of the context.
It may therefore favor text that is easier for the language model to predict rather than text that provides stronger evidence for the query.
This motivates reverse query evidence as our evidence-first criterion.
Standalone evidence, however, cannot capture interactions induced by sequential compression.
Redundancy, complementarity, information preemption, and the target budget can make the best ordering depend on the full context collection.
We therefore use $E_\theta(c_i;q)$ to construct an evidence-first anchor, while applying the same reverse query-evidence principle to the content retained after compression when evaluating alternative candidate orderings.
Section~\ref{sec:morse} instantiates these two uses within \texttt{MORSE}.

\section{Proposed Method: \texttt{MORSE}}
\label{sec:morse}

\subsection{Evidence-First Anchor}
\label{sec:reverse_likelihood}

Following the reverse query-evidence principle introduced in Section~\ref{sec:evidence_first},
\texttt{MORSE} first evaluates each retrieved context independently:
\begin{equation}
    r(c_i;q)
    =
    E_\theta(c_i;q)
    =
    L_\theta(q)
    -
    L_\theta(q\mid c_i).
    \label{eq:reverse_score}
\end{equation}
For a fixed query, this score induces the same ordering as reverse query likelihood $\log p_\theta(q\mid c_i)$.
We sort the contexts in decreasing order of $r(c_i;q)$ to obtain the reverse permutation
$\pi_0=(\pi_1,\ldots,\pi_n)$, such that
$r(c_{\pi_1};q)\geq\cdots\geq r(c_{\pi_n};q)$.
This provides an evidence-first anchor that prioritizes contexts with stronger standalone query evidence.

\subsection{Compression-Aware Global Exploration}
\label{sec:global_exploration}

Standalone evidence does not capture the interactions induced by sequential compression.
\texttt{MORSE} therefore combines the reverse anchor $\pi_0$ with up to $K-1$ additional unique global random permutations of the same contexts, sampled without replacement and excluding $\pi_0$.
Let $\mathcal{R}_{K-1}(\mathcal{C};\pi_0)$ denote this set.
The complete candidate set is
\begin{equation}
    \Pi_K(\pi_0)
    =
    \{\pi_0\}
    \cup
    \mathcal{R}_{K-1}(\mathcal{C};\pi_0),
    \label{eq:morse_candidates}
\end{equation}
with $|\Pi_K(\pi_0)|=\min\{K,n!\}$ when sufficient distinct permutations are available.

For each candidate permutation $\pi\in\Pi_K(\pi_0)$,
the contexts are serialized according to $\pi$ and processed by the same sequential compressor under token budget $B$:
\begin{equation}
    \widetilde{\mathcal{C}}_{\pi,B}
    =
    \operatorname{Compress}(q,\mathcal{C}_\pi,B).
    \label{eq:morse_compress}
\end{equation}
The retained units are then restored to their original document and within-document order:
\begin{equation}
    \widehat{\mathcal{C}}_{\pi,B}
    =
    \operatorname{Canon}
    \left(
        \widetilde{\mathcal{C}}_{\pi,B}
    \right).
    \label{eq:morse_canonicalize}
\end{equation}
Canonicalization ensures that candidate selection evaluates differences in retained content rather than differences in downstream presentation order.

We then apply the same reverse query-evidence principle used for the evidence-first anchor to the content that actually survives compression:
\begin{equation}
    J_B(\pi)
    =
    E_\theta
    \left(
        \widehat{\mathcal{C}}_{\pi,B};q
    \right)
    =
    L_\theta(q)
    -
    L_\theta
    \left(
        q
        \mid
        \widehat{\mathcal{C}}_{\pi,B}
    \right).
    \label{eq:compression_objective}
\end{equation}
Thus, $r(c_i;q)$ measures standalone query evidence before interactions among contexts are realized, whereas $J_B(\pi)$ measures the query evidence retained after the candidate ordering has acted through the compressor and target budget.

The final candidate is selected as
\begin{equation}
    \pi_{\texttt{MORSE}}
    =
    \arg\max_{\pi\in\Pi_K(\pi_0)}
    J_B(\pi).
    \label{eq:morse_selection}
\end{equation}
Because $\pi_0$ is always included in the candidate set, the evidence-first anchor remains available whenever none of the explored alternatives achieves a higher retained-evidence score.
The compressed output associated with $\pi_{\texttt{MORSE}}$ is used as the final result.
Unless otherwise stated, we use $K=5$.

\section{Experiments}
\label{sec:experiments}


\subsection{Experimental Setup}
\label{sec:experimental_setup}

\textbf{Datasets.}
We evaluate on HotpotQA \citep{yang2018hotpotqa} and 2WikiMultiHopQA \citep{ho2020constructing} under their native context collections and a larger \textsc{True-20} stress-test setting that appends non-supporting distractors while preserving the original questions, answers, and supporting evidence.
We use $500$ HotpotQA examples for the primary native \textsc{1P} evaluation and fixed $100$-example subsets for 2WikiMultiHopQA, \textsc{True-20}, and all \textsc{ISD} experiments.
Details are provided in Appendix~\ref{app:experimental_setup}.


\textbf{Baselines.}
We compare \texttt{MORSE} with \textsc{Independent}, five static baselines (\textsc{Original}, \textsc{Random-5}, \textsc{Length}, \textsc{Forward}, and \textsc{Reverse}), and compute-matched \textsc{RandomSearch-5}, which evaluates five random permutations with the same objective $J_B$.
Both search methods are averaged over seeds 42--46; Appendix~\ref{app:baselines} provides full definitions.

\textbf{Compression Setting.}
Our primary compressor, \textsc{1P}, computes sentence-level sequential chain-QMI scores once per ordering, while \textsc{ISD} serves as an iterative robustness test that recomputes affected scores after each sentence deletion.
Both use the same Qwen2.5-0.5B-Instruct likelihood scorer, sentence-level units, exact token-budget accounting, and $4\times$, $8\times$, and $16\times$ compression.
Implementation details are provided in Appendix~\ref{app:experimental_setup}.
\textbf{Evaluation Metrics.}
We report Supporting-Fact Recall (SF-R) as the primary measure of compression-stage evidence preservation, with Answer F1/EM serving as downstream QA metrics that also reflect reader and presentation effects.
For \textsc{1P} \citep{jiang2023llmlingua}, retained content is restored to canonical order before QA; for \textsc{ISD} \citep{jiang2024longllmlingua}, all methods use the same post-compression reverse reranking so that compression order and downstream presentation are separated.
Appendix~\ref{app:experimental_setup} provides the complete evaluation protocol.


\begin{table*}[t]
    \centering
    \caption{
        \textbf{Evidence retention and downstream QA across datasets, context settings, and compression procedures.}
        \textsc{1P} denotes the primary one-pass sequential chain-QMI compressor, and
        \textsc{ISD} denotes Iterative Sentence Deletion.
        SF-R denotes Supporting-Fact Recall.
        Results are reported under $4\times$, $8\times$, and $16\times$ compression.
        \textsc{MORSE} uses the practical $K=5$ search budget, evaluating the Reverse anchor together with four unique random permutations and selecting the candidate maximizing $J_B$.
        \textsc{RandomSearch-5} is the compute-matched search baseline that evaluates five unique global random permutations and selects the candidate maximizing $J_B$.
        For each metric, the best result is shown in \textbf{bold} and the runner-up is \underline{underlined}.
        Rankings use the underlying unrounded scores before rounding to two decimals; exact unrounded ties are deterministically broken by table order so each cell has a unique best and runner-up.
        Full-precision results are reported in Appendix~\ref{app:table1_detailed}.
        Higher is better.
    }
    \label{tab:main_full_results}

    \fontsize{6.7}{7.5}\selectfont
    \setlength{\tabcolsep}{1.45pt}
    \renewcommand{\arraystretch}{0.92}


    \textbf{(a) HotpotQA}

    \vspace{2pt}

    \begin{tabular}{
        @{}ll
        ccc ccc ccc
        @{\hspace{5pt}}
        ccc ccc ccc@{}
    }
        \toprule

        \multirow{3}{*}{\textbf{Method}}
        & \multirow{3}{*}{\textbf{Proc.}}
        & \multicolumn{9}{c}{\textbf{Native}}
        & \multicolumn{9}{c}{\textbf{\textsc{True-20}}}
        \\

        \cmidrule(lr){3-11}
        \cmidrule(lr){12-20}

        &
        & \multicolumn{3}{c}{$4\times$}
        & \multicolumn{3}{c}{$8\times$}
        & \multicolumn{3}{c}{$16\times$}
        & \multicolumn{3}{c}{$4\times$}
        & \multicolumn{3}{c}{$8\times$}
        & \multicolumn{3}{c}{$16\times$}
        \\

        \cmidrule(lr){3-5}
        \cmidrule(lr){6-8}
        \cmidrule(lr){9-11}
        \cmidrule(lr){12-14}
        \cmidrule(lr){15-17}
        \cmidrule(lr){18-20}

        &
        & SF-R & F1 & EM
        & SF-R & F1 & EM
        & SF-R & F1 & EM
        & SF-R & F1 & EM
        & SF-R & F1 & EM
        & SF-R & F1 & EM
        \\

        \midrule
        Independent
        & \textsc{1P}
        & 0.54 & 0.54 & 0.43
        & 0.37 & 0.48 & 0.38
        & 0.24 & 0.38 & 0.30
        & 0.68 & 0.64 & 0.54
        & 0.47 & 0.50 & 0.45
        & 0.28 & 0.41 & 0.36
        \\

        & \textsc{ISD}
        & 0.53 & 0.61 & 0.52
        & 0.31 & 0.44 & 0.38
        & 0.20 & 0.41 & 0.34
        & 0.68 & 0.65 & 0.54
        & 0.47 & 0.48 & 0.42
        & 0.28 & 0.44 & 0.38
        \\

\addlinespace[1pt]
        Original
        & \textsc{1P}
        & 0.59 & 0.55 & 0.43
        & 0.44 & 0.51 & 0.39
        & 0.29 & 0.41 & 0.32
        & 0.58 & 0.56 & 0.47
        & 0.50 & 0.50 & 0.42
        & 0.39 & 0.50 & 0.42
        \\

        & \textsc{ISD}
        & 0.56 & 0.59 & 0.51
        & 0.38 & 0.51 & 0.43
        & 0.21 & 0.41 & 0.34
        & 0.66 & 0.64 & 0.53
        & 0.55 & 0.59 & 0.51
        & 0.39 & 0.49 & 0.41
        \\

\addlinespace[1pt]
        Random-5
        & \textsc{1P}
        & 0.61 & 0.56 & 0.44
        & 0.45 & 0.49 & 0.38
        & 0.30 & 0.41 & 0.33
        & 0.67 & 0.61 & 0.51
        & 0.53 & 0.57 & 0.49
        & 0.40 & 0.49 & 0.41
        \\

        & \textsc{ISD}
        & 0.57 & 0.56 & 0.47
        & 0.38 & 0.49 & 0.42
        & 0.23 & 0.42 & 0.36
        & 0.70 & 0.64 & 0.54
        & 0.53 & 0.58 & 0.50
        & 0.39 & 0.51 & 0.42
        \\

\addlinespace[1pt]
        Length
        & \textsc{1P}
        & 0.66 & 0.60 & 0.47
        & 0.51 & 0.54 & 0.44
        & 0.35 & 0.45 & 0.35
        & 0.66 & 0.62 & 0.52
        & 0.56 & \underline{0.60} & 0.51
        & 0.46 & \underline{0.58} & \textbf{0.50}
        \\

        & \textsc{ISD}
        & 0.71 & \underline{0.64} & \underline{0.54}
        & 0.49 & 0.58 & \underline{0.52}
        & 0.29 & 0.51 & \textbf{0.44}
        & 0.75 & \textbf{0.69} & \textbf{0.60}
        & 0.61 & \underline{0.63} & \underline{0.55}
        & 0.49 & 0.57 & \underline{0.51}
        \\

\addlinespace[1pt]
        Forward
        & \textsc{1P}
        & 0.66 & 0.62 & 0.49
        & 0.52 & 0.55 & \underline{0.46}
        & 0.37 & 0.47 & 0.37
        & 0.68 & 0.64 & \underline{0.55}
        & 0.59 & 0.58 & \textbf{0.52}
        & 0.47 & 0.54 & 0.47
        \\

        & \textsc{ISD}
        & \underline{0.71} & \textbf{0.68} & \textbf{0.61}
        & \underline{0.54} & \textbf{0.63} & \textbf{0.55}
        & 0.32 & 0.52 & 0.43
        & 0.78 & \underline{0.69} & \underline{0.59}
        & \underline{0.66} & \textbf{0.66} & \textbf{0.60}
        & 0.53 & \textbf{0.61} & \textbf{0.55}
        \\

\addlinespace[1pt]
        Reverse
        & \textsc{1P}
        & 0.71 & 0.62 & 0.49
        & 0.58 & 0.56 & 0.44
        & 0.41 & 0.48 & 0.37
        & 0.66 & 0.60 & 0.50
        & 0.57 & 0.59 & 0.50
        & 0.51 & 0.56 & 0.47
        \\

        & \textsc{ISD}
        & 0.70 & 0.63 & 0.54
        & 0.54 & 0.58 & 0.49
        & 0.34 & 0.52 & 0.42
        & \underline{0.78} & 0.67 & 0.57
        & 0.65 & 0.59 & 0.51
        & \underline{0.55} & 0.56 & 0.49
        \\

\addlinespace[1pt]
        RandomSearch-5
        & \textsc{1P}
        & \underline{0.72} & \underline{0.62} & \textbf{0.50}
        & \underline{0.58} & \underline{0.57} & 0.45
        & \underline{0.41} & \underline{0.48} & \underline{0.38}
        & \underline{0.76} & \underline{0.66} & 0.55
        & \underline{0.65} & 0.58 & 0.49
        & \underline{0.52} & 0.56 & 0.46
        \\

        & \textsc{ISD}
        & 0.69 & 0.62 & 0.52
        & 0.52 & 0.58 & 0.49
        & \underline{0.35} & \underline{0.52} & \underline{0.44}
        & 0.77 & 0.67 & 0.57
        & 0.63 & 0.62 & 0.52
        & 0.50 & 0.57 & 0.49
        \\

\midrule

        \textbf{\texttt{MORSE}}
        & \textsc{1P}
        & \textbf{0.74} & \textbf{0.62} & \underline{0.50}
        & \textbf{0.62} & \textbf{0.59} & \textbf{0.46}
        & \textbf{0.44} & \textbf{0.50} & \textbf{0.39}
        & \textbf{0.78} & \textbf{0.67} & \textbf{0.57}
        & \textbf{0.68} & \textbf{0.61} & \underline{0.52}
        & \textbf{0.58} & \textbf{0.58} & \underline{0.48}
        \\

        & \textsc{ISD}
        & \textbf{0.72} & 0.64 & 0.54
        & \textbf{0.57} & \underline{0.62} & 0.51
        & \textbf{0.37} & \textbf{0.53} & 0.44
        & \textbf{0.80} & 0.68 & 0.58
        & \textbf{0.68} & 0.62 & 0.51
        & \textbf{0.57} & \underline{0.59} & 0.48
        \\
        \bottomrule
    \end{tabular}

    \vspace{8pt}


    \textbf{(b) 2WikiMultiHopQA}

    \vspace{2pt}

    \begin{tabular}{
        @{}ll
        ccc ccc ccc
        @{\hspace{5pt}}
        ccc ccc ccc@{}
    }
        \toprule

        \multirow{3}{*}{\textbf{Method}}
        & \multirow{3}{*}{\textbf{Proc.}}
        & \multicolumn{9}{c}{\textbf{Native}}
        & \multicolumn{9}{c}{\textbf{\textsc{True-20}}}
        \\

        \cmidrule(lr){3-11}
        \cmidrule(lr){12-20}

        &
        & \multicolumn{3}{c}{$4\times$}
        & \multicolumn{3}{c}{$8\times$}
        & \multicolumn{3}{c}{$16\times$}
        & \multicolumn{3}{c}{$4\times$}
        & \multicolumn{3}{c}{$8\times$}
        & \multicolumn{3}{c}{$16\times$}
        \\

        \cmidrule(lr){3-5}
        \cmidrule(lr){6-8}
        \cmidrule(lr){9-11}
        \cmidrule(lr){12-14}
        \cmidrule(lr){15-17}
        \cmidrule(lr){18-20}

        &
        & SF-R & F1 & EM
        & SF-R & F1 & EM
        & SF-R & F1 & EM
        & SF-R & F1 & EM
        & SF-R & F1 & EM
        & SF-R & F1 & EM
        \\

        \midrule
        Independent
        & \textsc{1P}
        & 0.36 & 0.45 & 0.42
        & 0.16 & 0.36 & 0.34
        & 0.08 & 0.33 & 0.31
        & 0.48 & 0.45 & 0.43
        & 0.27 & 0.38 & 0.37
        & 0.12 & 0.29 & 0.27
        \\

        & \textsc{ISD}
        & 0.36 & \textbf{0.47} & \textbf{0.42}
        & 0.16 & 0.35 & 0.32
        & 0.08 & 0.31 & 0.29
        & 0.48 & 0.47 & 0.42
        & 0.27 & 0.40 & 0.37
        & 0.12 & 0.33 & 0.30
        \\

\addlinespace[1pt]
        Original
        & \textsc{1P}
        & 0.46 & 0.44 & 0.42
        & 0.25 & 0.37 & 0.33
        & 0.13 & 0.29 & 0.25
        & 0.53 & 0.51 & \underline{0.46}
        & 0.40 & 0.36 & 0.31
        & 0.22 & 0.35 & 0.31
        \\

        & \textsc{ISD}
        & 0.41 & 0.44 & 0.39
        & 0.22 & 0.37 & 0.33
        & 0.09 & 0.35 & 0.31
        & 0.59 & 0.47 & 0.41
        & 0.38 & 0.41 & 0.35
        & 0.20 & 0.37 & 0.32
        \\

\addlinespace[1pt]
        Random-5
        & \textsc{1P}
        & 0.45 & 0.43 & 0.39
        & 0.24 & 0.37 & 0.33
        & 0.14 & 0.32 & 0.27
        & 0.58 & 0.47 & 0.42
        & 0.41 & 0.42 & 0.37
        & 0.23 & 0.34 & 0.30
        \\

        & \textsc{ISD}
        & 0.41 & 0.43 & 0.37
        & 0.22 & 0.37 & 0.33
        & 0.08 & 0.30 & 0.26
        & 0.55 & 0.45 & 0.39
        & 0.33 & 0.39 & 0.35
        & 0.15 & 0.31 & 0.28
        \\

\addlinespace[1pt]
        Length
        & \textsc{1P}
        & 0.40 & 0.40 & 0.36
        & 0.22 & 0.40 & 0.34
        & 0.12 & 0.30 & 0.25
        & 0.61 & \underline{0.52} & 0.46
        & 0.35 & 0.41 & 0.35
        & 0.20 & 0.35 & 0.31
        \\

        & \textsc{ISD}
        & 0.36 & 0.38 & 0.35
        & 0.19 & 0.39 & 0.34
        & 0.07 & 0.33 & 0.27
        & 0.52 & 0.46 & 0.40
        & 0.34 & 0.37 & 0.32
        & 0.14 & 0.35 & 0.30
        \\

\addlinespace[1pt]
        Forward
        & \textsc{1P}
        & 0.42 & 0.43 & 0.39
        & 0.22 & 0.34 & 0.29
        & 0.12 & 0.35 & 0.29
        & 0.60 & 0.46 & 0.41
        & 0.38 & 0.42 & 0.36
        & 0.21 & 0.33 & 0.26
        \\

        & \textsc{ISD}
        & 0.39 & 0.35 & 0.32
        & 0.18 & 0.36 & 0.31
        & 0.06 & 0.33 & 0.27
        & 0.53 & 0.44 & 0.37
        & 0.32 & 0.38 & 0.35
        & 0.13 & 0.31 & 0.26
        \\

\addlinespace[1pt]
        Reverse
        & \textsc{1P}
        & \underline{0.55} & 0.47 & 0.43
        & \underline{0.35} & \underline{0.41} & \underline{0.38}
        & 0.21 & \underline{0.38} & 0.33
        & 0.59 & 0.48 & 0.44
        & 0.47 & 0.48 & 0.44
        & 0.32 & 0.37 & 0.32
        \\

        & \textsc{ISD}
        & \underline{0.55} & 0.39 & 0.35
        & \underline{0.31} & 0.37 & 0.32
        & 0.16 & \underline{0.38} & \underline{0.34}
        & \underline{0.64} & \textbf{0.50} & \textbf{0.43}
        & \underline{0.50} & 0.42 & 0.37
        & 0.25 & 0.35 & 0.31
        \\

\addlinespace[1pt]
        RandomSearch-5
        & \textsc{1P}
        & 0.55 & \underline{0.48} & \underline{0.43}
        & 0.35 & 0.40 & 0.36
        & \underline{0.23} & 0.38 & \underline{0.34}
        & \underline{0.64} & \textbf{0.52} & \textbf{0.47}
        & \underline{0.47} & \underline{0.48} & \underline{0.44}
        & \underline{0.33} & \underline{0.42} & \textbf{0.38}
        \\

        & \textsc{ISD}
        & 0.52 & \underline{0.45} & \underline{0.40}
        & 0.29 & \underline{0.40} & \underline{0.35}
        & \underline{0.17} & 0.37 & 0.33
        & 0.62 & 0.48 & 0.43
        & 0.44 & \underline{0.43} & \underline{0.39}
        & \underline{0.26} & \underline{0.39} & \underline{0.36}
        \\

\midrule

        \textbf{\texttt{MORSE}}
        & \textsc{1P}
        & \textbf{0.59} & \textbf{0.49} & \textbf{0.44}
        & \textbf{0.37} & \textbf{0.42} & \textbf{0.38}
        & \textbf{0.24} & \textbf{0.40} & \textbf{0.35}
        & \textbf{0.66} & 0.51 & 0.45
        & \textbf{0.51} & \textbf{0.51} & \textbf{0.47}
        & \textbf{0.35} & \textbf{0.42} & \underline{0.37}
        \\

        & \textsc{ISD}
        & \textbf{0.56} & 0.44 & 0.39
        & \textbf{0.34} & \textbf{0.41} & \textbf{0.36}
        & \textbf{0.19} & \textbf{0.39} & \textbf{0.35}
        & \textbf{0.64} & \underline{0.48} & \underline{0.43}
        & \textbf{0.52} & \textbf{0.45} & \textbf{0.40}
        & \textbf{0.31} & \textbf{0.40} & \textbf{0.36}
        \\
        \bottomrule
    \end{tabular}

\end{table*}


\subsection{Numerical Result Analysis}
\label{sec}
Table~\ref{tab:main_full_results} shows that \textsc{Reverse} is generally the strongest static ordering baseline, supporting reverse query likelihood as an effective evidence-first anchor.
More importantly, \texttt{MORSE}-5 consistently outperforms the compute-matched \textsc{RandomSearch-5} baseline in evidence retention: SF-R improves in all 12 \textsc{1P} settings and all 12 \textsc{ISD} settings, indicating that the Reverse anchor provides a systematic advantage beyond generic compression-aware permutation search.
The gains remain pronounced in the larger \textsc{True-20} settings, where additional distractors increase competition for the compression budget.
Relative to \textsc{Reverse} alone, \texttt{MORSE}-5 further improves SF-R across all settings, confirming that compression-aware candidate selection provides complementary gains beyond evidence-first anchor.
The downstream results are positive but less uniform.
Under \textsc{1P}, improvements in evidence retention are accompanied by broadly consistent gains in Answer F1 and EM, whereas under \textsc{ISD} the corresponding downstream gains vary more across datasets and compression ratios.
In particular, higher SF-R does not always translate into higher F1 or EM under milder compression.
We therefore interpret SF-R as a direct measure of annotated evidence preservation rather than as a complete surrogate for downstream QA performance.





\subsection{Search-Budget Ablation}
\label{sec:search_budget}

\begin{table*}[t]    
\centering    
\caption{    
\textbf{Quality--efficiency tradeoff of the \texttt{MORSE} candidate budget on HotpotQA.}    
We vary the number of candidate orderings $K$, where $K=1$ corresponds to the Reverse ordering without search.    
For $K>1$, \texttt{MORSE} evaluates the Reverse anchor together with $K-1$ unique random permutations and selects the compressed output with the highest compression objective $J_B$.    
Latency is measured using the exact candidate-parallel implementation over five NVIDIA RTX A6000 GPUs; independent candidates are evaluated concurrently while preserving batch-dimension-one likelihood scoring.    
SF-R, Answer F1, and EM are evaluated on the fixed SAME-100 HotpotQA Native set under $4\times$, $8\times$, and $16\times$ compression.    
All quality results use the unified downstream evaluation protocol.    
}    
\label{tab:k_efficiency}    
\fontsize{7.4}{8.4}\selectfont    
\setlength{\tabcolsep}{3.5pt}    
\renewcommand{\arraystretch}{1.10}

\begin{tabular}{@{}ccc ccc ccc ccc@{}}        
\toprule        
\multirow{2}{*}{\textbf{$K$}}        
& \multirow{2}{*}{\textbf{Candidates}}        
& \multirow{2}{*}{\textbf{Latency (s/ex)}}        
& \multicolumn{3}{c}{\textbf{$4\times$}}        
& \multicolumn{3}{c}{\textbf{$8\times$}}        
& \multicolumn{3}{c}{\textbf{$16\times$}} \\        
\cmidrule(lr){4-6}        
\cmidrule(lr){7-9}        
\cmidrule(l){10-12}        
& & &        
\textbf{SF-R} & \textbf{F1} & \textbf{EM} &        
\textbf{SF-R} & \textbf{F1} & \textbf{EM} &        
\textbf{SF-R} & \textbf{F1} & \textbf{EM} \\        
\midrule

$1$ (Reverse)        
& 1        
& \textbf{0.210}        
& 0.673 & 0.616 & 0.530        
& 0.544 & 0.600 & 0.500        
& 0.386 & 0.509 & 0.390 \\

$5$ (\texttt{MORSE})        
& 5        
& 1.890        
& 0.718 & 0.636 & 0.536        
& 0.595 & 0.606 & 0.494        
& 0.426 & \textbf{0.512} & 0.406 \\

$10$        
& 10        
& 3.600        
& \textbf{0.742} & \textbf{0.639} & \textbf{0.548}        
& \textbf{0.637} & \textbf{0.620} & \textbf{0.508}        
& \textbf{0.444} & 0.505 & \textbf{0.410} \\

\bottomrule    
\end{tabular}
\end{table*}

Increasing $K$ improves evidence retention by exploring a broader set of compression-aware orderings, with diminishing returns as the search budget grows.
Moving from $K=1$ to a small multi-candidate search captures much of the benefit, while larger $K$ mainly provides incremental improvements at higher cost.
We therefore use $K=5$ as the default practical operating point, balancing search quality and latency, while larger $K$ remains useful when additional test-time computation is available.

\subsection{Selection-Objective Ablation}
\label{sec:jb_ablation}

We isolate the candidate-selection objective by fixing all candidate permutations and compressed outputs and varying only the selection score.
The MORSE pool contains $\{\text{Reverse},R_1,\ldots,R_4\}$, while the random pool contains $\{R_1,\ldots,R_5\}$.
Besides the reverse objective $J_B$, we evaluate a forward selector,
$J_B^{\mathrm{fwd}}(\pi)=\log p_\theta(\widehat{\mathcal C}_{\pi,B}\mid q)$,
where $\widehat{\mathcal C}_{\pi,B}$ is the canonically reordered compressed output.
Forward-Norm divides this score by the number of scored output tokens to control for length.
As shown in Table~\ref{tab:jb_ablation}, $J_B$ selects substantially higher-SF-R candidates and correlates more strongly with SF-R than either forward alternative.
Full results and controls are provided in Appendix~\ref{app:jb_ablation}.

\begin{table}[t]
\centering
\caption{
\textbf{Ablation of the compression-aware selection objective.}
Selected SF-R and within-example Spearman correlation are averaged equally over the 12 1P settings.
}
\label{tab:jb_ablation}
\small
\setlength{\tabcolsep}{5.0pt}
\renewcommand{\arraystretch}{1.08}
\begin{tabular}{@{}lccc@{}}
\toprule
\textbf{Selection score}
& \textbf{MORSE pool}
& \textbf{Random pool}
& $\boldsymbol{\rho}$\textbf{ w/ SF-R} \\
\midrule
Forward
& 0.490 & 0.465 & 0.230 \\
Forward-Norm.
& 0.493 & 0.470 & 0.251 \\
Reverse $J_B$
& \textbf{0.547} & \textbf{0.518} & \textbf{0.401} \\
\bottomrule
\end{tabular}
\end{table}

\subsection{Generalization}
\label{sec:generalization}

We further evaluate MORSE under a $2\times2$ scorer--dataset design using Qwen2.5-0.5B and OLMo-2-1B \citep{olmo20242olmo2furious} as compression scorers, and HotpotQA TRUE-20 and MuSiQue \citep{trivedi2022musique} as datasets.
All settings use 20 contexts, 1P compression, $K=5$, and the same Qwen2.5-7B downstream reader.
For MuSiQue, we report Supporting-Paragraph Recall (SP-R) because its gold evidence annotations are at the paragraph level.
MORSE outperforms Reverse in all four scorer--dataset combinations and remains better than compute-matched RandomSearch-5 on average in every setting.
The gain over RandomSearch-5 transfers clearly under either the scorer or dataset shift individually, but becomes small and statistically inconclusive under the joint OLMo--MuSiQue shift.
Appendix~\ref {app:generalization} provides full ratio-specific retention, QA results, confidence intervals, and MuSiQue diagnostics.

\begin{table}[t]
\centering
\caption{
\textbf{Cross-model and cross-dataset generalization.}
Retention is SF-R for HotpotQA and SP-R for MuSiQue, averaged equally over $4\times$, $8\times$, and $16\times$ compression.
}
\label{tab:generalization_main}
\small
\setlength{\tabcolsep}{6.0pt}
\renewcommand{\arraystretch}{1.08}
\begin{tabular}{@{}llccc@{}}
\toprule
\textbf{Scorer}
& \textbf{Dataset}
& \textbf{Reverse}
& \textbf{RandomSearch-5}
& \textbf{MORSE-5} \\
\midrule
Qwen & HotpotQA & 0.580 & 0.646 & \textbf{0.679} \\
OLMo & HotpotQA & 0.462 & 0.491 & \textbf{0.535} \\
Qwen & MuSiQue & 0.654 & 0.692 & \textbf{0.713} \\
OLMo & MuSiQue & 0.509 & 0.618 & \textbf{0.626} \\
\bottomrule
\end{tabular}
\end{table}

\begin{figure*}[!t]
\centering

\begin{subfigure}[t]{0.315\textwidth}
\centering
\includegraphics[width=\linewidth]{
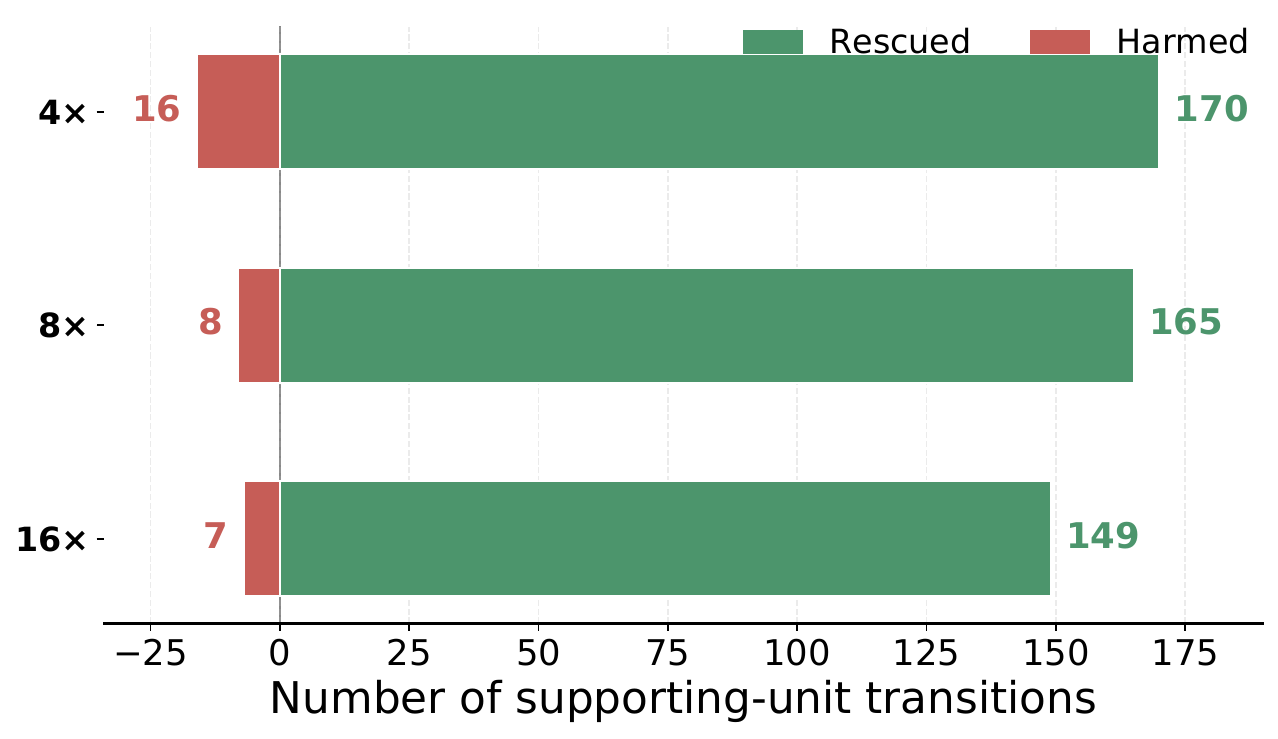
}
\caption{
Evidence-first support transitions.
}
\label{fig:pair_swap_rescue}
\end{subfigure}
\hfill
\begin{subfigure}[t]{0.315\textwidth}
\centering
\includegraphics[width=\linewidth]{
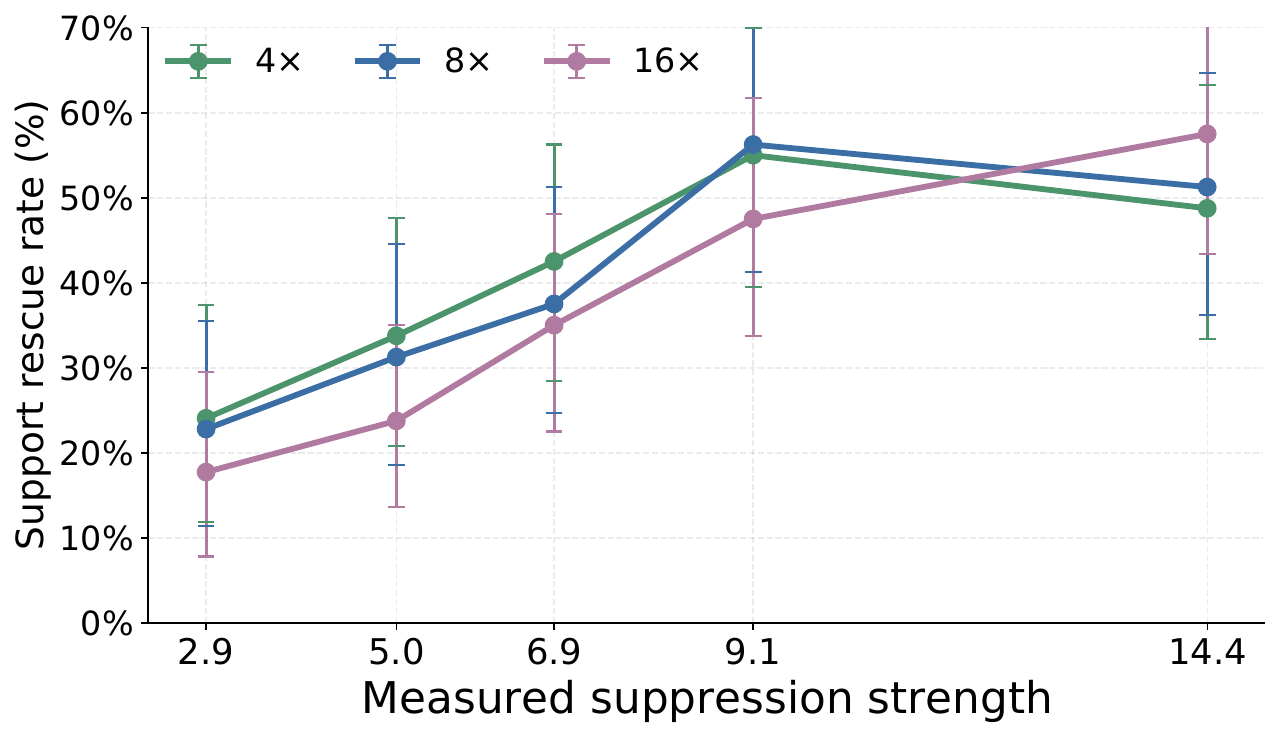
}
\caption{
Rescue rate by suppression strength.
}
\label{fig:pair_swap_strength}
\end{subfigure}
\hfill
\begin{subfigure}[t]{0.315\textwidth}
\centering
\includegraphics[width=\linewidth]{
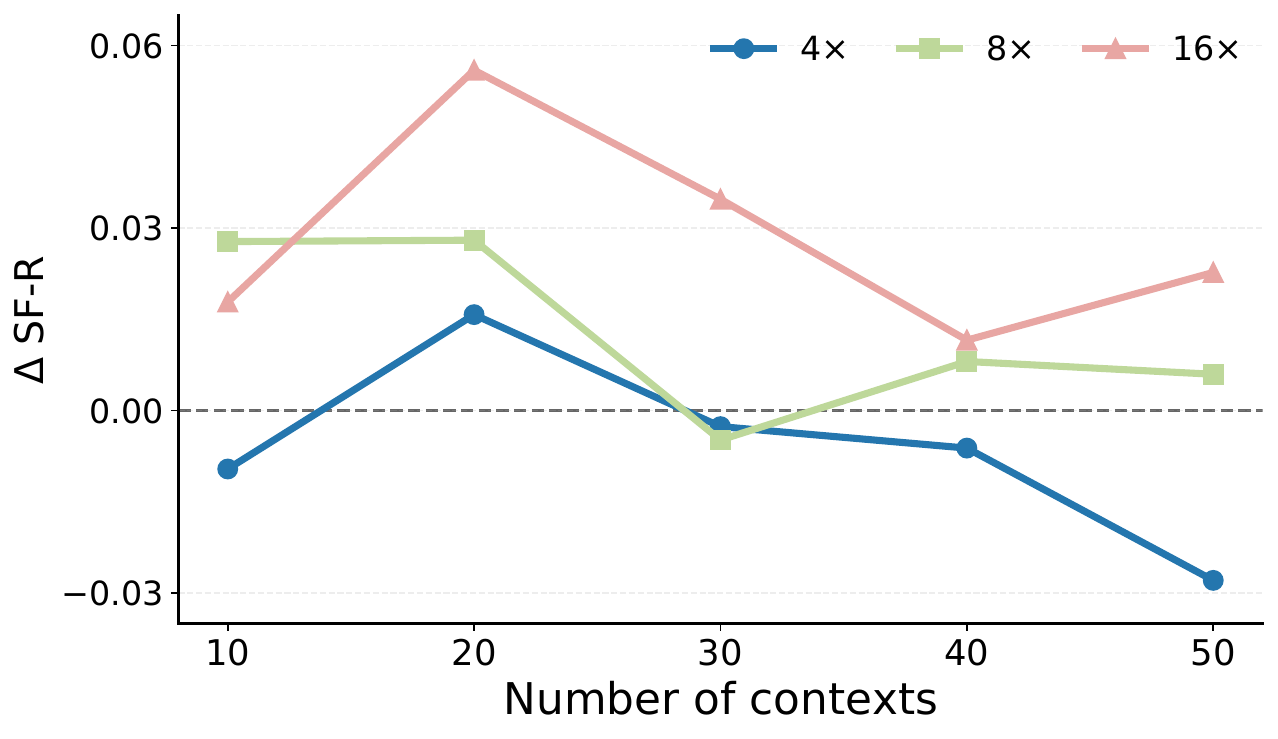
}
\caption{
MORSE--RandomSearch gap with context count.
}
\label{fig:context_scaling}
\end{subfigure}

\vspace{-1mm}

\caption{
\textbf{Information preemption and context-count scaling.}
\textbf{(a)} Placing evidence before its measured preemptor rescues far more supporting units than it harms: 170 vs.\ 16 at $4\times$, 165 vs.\ 8 at $8\times$, and 149 vs.\ 7 at $16\times$.
\textbf{(b)} Stronger measured suppression is associated with a higher probability that the swap rescues supporting evidence; error bars denote clustered-bootstrap 95\% confidence intervals.
\textbf{(c)} The SF-R difference between \texttt{MORSE-5} and compute-matched \textsc{RandomSearch-5} varies with context count and compression regime, with clearer gains under stronger compression.
}
\label{fig:pair_swap_mechanism}

\vspace{-2mm}
\end{figure*}

\paragraph{Scaling to larger context collections.}
We further evaluate context-count scaling on the fixed SAME-100 HotpotQA population using nested context collections of 10, 20, 30, 40, and 50 contexts under 1P compression.
As shown in Figure~\ref{fig:context_scaling}, \texttt{MORSE-5} remains substantially stronger than static Reverse ordering across all 15 context-count--compression settings, with every paired 95\% confidence interval above zero.
Relative to compute-matched \textsc{RandomSearch-5}, the additional benefit of the Reverse anchor is more regime-dependent: the two methods are often statistically comparable as the collection grows, while clearer MORSE gains appear under stronger compression.
Because the nominal compression ratio is fixed, the absolute retained-token budget grows with input size; we therefore interpret this experiment as a scaling stress test rather than a fixed-budget intervention.
Full SF-R, downstream QA, and paired confidence intervals are reported in Appendix~\ref{app:context_scaling}.


\subsection{Compression Order versus Presentation Order}
\label{sec:compression_vs_presentation}
Reverse ordering exhibits a clear \emph{compression-stage} effect. When we change only the compression order while keeping downstream presentation fixed, Reverse improves SF-R in all 12 settings and improves F1 and EM in 11 of 12 settings, with average gains of 9.58, 5.38, and 4.77 points, respectively. In contrast, changing only the downstream presentation order while holding the compressed content fixed has a negligible and inconsistent effect, yielding average gains of only 0.05 F1 and 0.17 EM points. This contrast indicates that the primary role of relevance-based reordering in our sequential compression pipeline is to change which evidence survives compression, rather than simply placing surviving evidence in more favorable positions for the downstream reader. More broadly, these results suggest an additional mechanism behind relevance-based document reordering in prior sequential compression methods such as LongLLMLingua: besides mitigating downstream positional bias, reordering may also improve downstream performance by changing evidence survival during compression itself~\citep{jiang2024longllmlingua}.

\begin{table}[t]
\centering
\caption{
\textbf{Decomposing the effect of reverse ordering into compression-stage and presentation-stage contributions.}
Results are averaged over all 12 settings
(HotpotQA / 2Wiki $\times$ Native / \textsc{True-20} $\times$ $4\times$ / $8\times$ / $16\times$).
Compression-only changes the order used during compression while keeping canonical downstream presentation.
Presentation-only changes only the downstream presentation order while keeping the compressed content fixed.
Combined changes both.
Win rates denote the fraction of settings with a positive improvement.
}
\label{tab:decomposition_summary}
\fontsize{8.0}{9.2}\selectfont
\setlength{\tabcolsep}{3.8pt}
\renewcommand{\arraystretch}{1.16}

\begin{tabular}{@{}lcccccc@{}}
\toprule
\multirow{2}{*}{\textbf{Effect}}
& \multicolumn{3}{c}{\textbf{Mean Improvement}}
& \multicolumn{3}{c}{\textbf{Win Rate}} \\
\cmidrule(lr){2-4}
\cmidrule(l){5-7}
& $\Delta$ SF-R
& $\Delta$ F1
& $\Delta$ EM
& SF-R
& F1
& EM \\
\midrule

Compression-only
& \textbf{+0.0958}
& \textbf{+0.0538}
& \textbf{+0.0477}
& \textbf{12/12}
& \textbf{11/12}
& \textbf{11/12}
\\

Presentation-only
& 0.0000
& +0.0005
& +0.0017
& 0/12
& 7/12
& 9/12
\\

Combined
& +0.0958
& +0.0403
& +0.0360
& \textbf{12/12}
& 9/12
& 9/12
\\

\bottomrule
\end{tabular}
\end{table}

\subsection{Controlled Pair-Swap Intervention}
\label{sec:pair_swap}

To directly test information preemption, we perform a controlled pair-swap intervention.
For each evidence context $c_e$ with gold supporting units 
\begin{equation}
\Delta_{p\rightarrow e}
=
\frac{1}{|\mathcal{G}_e|}
\sum_{u\in\mathcal{G}_e}
\left[s_0(u)-s_{p\prec e}(u)\right],
\label{eq:preemption_pair}
\end{equation}
where $s_0(u)$ is the score without $c_p$ preceding $c_e$, and $s_{p\prec e}(u)$ is the score when $c_p$ appears first.
For each example, we select the strongest non-gold preemptor and compare the matched orders $c_p\prec c_e$ and $c_e\prec c_p$, while keeping all other contexts, the compressor, and the budget fixed.
This yields 399 eligible pairs.
Figure~\ref{fig:pair_swap_rescue} shows that placing evidence before its preemptor rescues far more supporting units than it harms: 170 vs.\ 16 at $4\times$, 165 vs.\ 8 at $8\times$, and 149 vs.\ 7 at $16\times$.
The corresponding mean SF-R gains are 17.39, 17.08, and 16.06 percentage points.
Figure~\ref{fig:pair_swap_strength} further shows that stronger measured suppression is associated with a higher rescue probability, rising from 24.1\% to 48.8\% at $4\times$, 22.8\% to 51.3\% at $8\times$, and 17.7\% to 57.5\% at $16\times$ from the lowest to highest suppression quintile.
These results directly support information preemption: earlier contexts can suppress the incremental credit of later evidence, making it more likely to be removed.
\textsc{MORSE-5} also corrects measured harmful preemptor--evidence relations more frequently than compute-matched \textsc{RandomSearch-5} ($65.6\%$ vs.\ $55.4\%$); full correction, preservation, and intervention-linked analyses are provided in Appendix~\ref{app:preemption_correction}.

\Needspace{9\baselineskip}
\section{Conclusion}
\label{sec:conclusion}
We study context ordering in query-conditioned sequential compression.
Changing only the context permutation can substantially alter which supporting evidence survives compression.
We attribute this sensitivity to order-dependent credit assignment and information preemption, supported by controlled pair-swap interventions.
\texttt{MORSE} combines a reverse-likelihood evidence-first anchor with compression-aware global exploration and selects the candidate retaining the most query evidence.
Thus, ordering can materially change what the compressor preserves.

\bibliographystyle{morse_preprint}
\bibliography{main}

\newpage
\appendix

\section{Additional Experimental Details}
\label{app:details}

\subsection{Full Experimental Setup}
\label{app:experimental_setup}

\paragraph{Datasets.}
We evaluate on two multi-hop question-answering benchmarks, HotpotQA and 2WikiMultiHopQA, both of which require combining evidence distributed across multiple documents.
For HotpotQA, we use the distractor validation set, where the native context collections contain at most 10 contexts, with an average of 9.9 contexts per example, and evaluate the primary native setting on 500 examples.
For 2WikiMultiHopQA, we use the official development split with fixed sentence segmentation, where every example contains exactly 10 native contexts, and evaluate a fixed subset of 100 examples.
To evaluate robustness under larger and noisier context collections, we additionally construct a \textsc{True-20} stress-test setting for both datasets.
For each example, we preserve all native contexts and deterministically append non-supporting distractor contexts sampled from other examples in the same dataset until the collection contains exactly 20 contexts with unique titles.
The question, gold answer, supporting-fact annotations, and all native contexts remain unchanged.
We evaluate \textsc{True-20} on fixed subsets of 100 examples for both datasets, with the compression budget computed from the enlarged input using the same nominal compression ratios as in the corresponding native setting.
We treat \textsc{True-20} as a stress test rather than a causal intervention on context count, since augmentation also changes distractor composition, total input length, and the corresponding ratio-matched absolute budgets.
For the iterative-compression experiments, we use fixed subsets of 100 examples in all four dataset--context settings; within each dataset, the native and \textsc{True-20} settings use the same target questions.

\paragraph{Baselines.}
We compare \texttt{MORSE} with one non-sequential baseline, five static ordering baselines, and one compute-matched search baseline.
\textsc{Independent} removes cross-context sequential conditioning.
The static ordering baselines are \textsc{Original}, \textsc{Random-5}, \textsc{Length}, \textsc{Forward}, and \textsc{Reverse}; all use the same underlying sequential compressor and token budget and differ only in the supplied context order.
\textsc{Reverse} is also the evidence-first anchor used by \texttt{MORSE}.
\textsc{RandomSearch-5} evaluates five unique global random permutations with the same compressor and selects the candidate maximizing $J_B$, matching the $K=5$ search budget of \texttt{MORSE} while omitting the Reverse anchor.
This comparison isolates whether the informed anchor improves finite-budget search beyond generic compression-aware permutation exploration.
Detailed definitions of all baselines are given in Appendix~\ref{app:baselines}.

\paragraph{Compression Setting.}
Our primary compressor, denoted \textsc{1P}, computes sentence-level sequential chain-QMI scores once under the chosen context order and uses these fixed scores for budget-constrained selection.
As a robustness test, we additionally use an iterative sentence-deletion procedure, denoted \textsc{ISD}.
\textsc{ISD} repeatedly recomputes sequential chain-QMI scores under the current surviving history, deletes the currently lowest-scoring complete sentence, updates the affected scores, and continues until the target budget is reached.
Thus, \textsc{1P} evaluates each sentence once per candidate ordering, whereas \textsc{ISD} updates scores after deletion.
The atomic deletion unit is always a complete sentence, preserving the $(\text{title}, \text{sentence id})$ identities required by supporting-fact evaluation.
For \texttt{MORSE}, every candidate permutation is evaluated using the corresponding compressor itself, so the search objective remains compression-aware under both \textsc{1P} and \textsc{ISD}.
Both procedures use Qwen2.5-0.5B-Instruct as the likelihood scorer, the same preprocessing and exact token-budget accounting, and compression ratios of $4\times$, $8\times$, and $16\times$; each context is truncated to at most 180 tokens before compression.
We intentionally use a lightweight scorer because \texttt{MORSE} repeatedly evaluates likelihoods across candidate orderings at test time; this model estimates relative information value and is separate from the larger downstream model used to generate answers.

\paragraph{Evaluation Metrics.}
We evaluate compressed outputs using three complementary metrics.
\textbf{Supporting-Fact Recall (SF-R)} measures the fraction of annotated supporting evidence preserved after compression.
To measure downstream utility, we provide the compressed contexts to Qwen2.5-7B-Instruct and report \textbf{Answer F1} and \textbf{Exact Match (EM)} using the corresponding dataset evaluation protocol.
The downstream model uses greedy decoding with the same frozen prompt and generation configuration across methods.
For the primary \textsc{1P} experiments, retained contexts and sentences are restored to their canonical document order before downstream inference.
For the \textsc{ISD} experiments, every method instead uses the same post-compression Reverse query-likelihood reranking policy before downstream inference, separating the ordering used during compression from the ordering presented to the downstream model.
We therefore use \textsc{ISD} as a robustness test of the ordering phenomenon rather than interpreting absolute downstream differences between \textsc{1P} and \textsc{ISD} as a pure compressor-only effect.
For \textsc{Random-5}, all reported metrics are averaged over the same five deterministic permutations.
For \textsc{RandomSearch-5}, the same compression-aware objective $J_B$ selects one candidate from five unique global random permutations for each example and budget.
For the stochastic $K=5$ search methods, the reported formal results are averaged over five deterministic seeds, 42--46.

\subsection{Compression Order versus Presentation Order}
\label{app:compression_presentation}

To separate the effect of context ordering during compression from its effect on downstream presentation, we conduct a $2\times2$ decomposition using \textsc{Original} and \textsc{Reverse} ordering.
We independently vary (i) the order used by the sequential compressor and (ii) the order in which the surviving contexts are presented to the downstream QA model.
Canonical presentation restores the retained contexts and sentences to their original document and within-document order, whereas Reverse presentation orders the surviving contexts according to their pre-compression reverse query-likelihood ranking.
Because presentation occurs after compression, SF-R depends only on the compression order.

\begin{table*}[t]
\centering
\caption{
\textbf{Separating compression-order and downstream-presentation effects.}
We independently vary the ordering used during compression and the ordering used to present the surviving contexts to the downstream model.
Canonical presentation restores the retained content to its original document order, whereas Reverse presentation follows the pre-compression reverse query-likelihood ranking.
SF-R is unaffected by downstream presentation and is therefore summarized separately as the change induced by switching the compression order from Original to Reverse.
}
\label{tab:compression_presentation}
\fontsize{7.2}{8.2}\selectfont
\setlength{\tabcolsep}{3.2pt}
\renewcommand{\arraystretch}{1.08}
\begin{tabular}{@{}llc cc cc cc cc@{}}
\toprule
\multirow{2}{*}{\textbf{Dataset}}
& \multirow{2}{*}{\textbf{Setting}}
& \multirow{2}{*}{\textbf{Ratio}}
& \multicolumn{2}{c}{\textbf{Original Compression}}
& \multicolumn{2}{c}{\textbf{Original Compression}}
& \multicolumn{2}{c}{\textbf{Reverse Compression}}
& \multicolumn{2}{c}{\textbf{Reverse Compression}} \\
\cmidrule(lr){4-5}
\cmidrule(lr){6-7}
\cmidrule(lr){8-9}
\cmidrule(l){10-11}
& &
& \multicolumn{2}{c}{Canonical Present.}
& \multicolumn{2}{c}{Reverse Present.}
& \multicolumn{2}{c}{Canonical Present.}
& \multicolumn{2}{c}{Reverse Present.} \\
& &
& F1 & EM
& F1 & EM
& F1 & EM
& F1 & EM \\
\midrule
HotpotQA & Native & $4\times$ & 0.551 & 0.434 & 0.560 & 0.438 & \textbf{0.618} & \textbf{0.492} & \textbf{0.621} & 0.486 \\
& & $8\times$ & 0.505 & 0.394 & 0.505 & 0.396 & \textbf{0.562} & \textbf{0.436} & 0.560 & \textbf{0.436} \\
& & $16\times$ & 0.414 & 0.318 & 0.416 & 0.322 & \textbf{0.479} & \textbf{0.370} & 0.472 & 0.356 \\
\addlinespace[2pt]
HotpotQA & \textsc{True-20} & $4\times$ & 0.565 & 0.470 & 0.600 & 0.500 & 0.597 & 0.500 & \textbf{0.618} & \textbf{0.530} \\
& & $8\times$ & 0.502 & 0.420 & 0.535 & 0.440 & \textbf{0.590} & 0.500 & 0.588 & \textbf{0.510} \\
& & $16\times$ & 0.505 & 0.420 & 0.506 & 0.430 & 0.563 & \textbf{0.470} & \textbf{0.564} & \textbf{0.470} \\
\addlinespace[2pt]
2Wiki & Native & $4\times$ & 0.438 & 0.420 & 0.410 & 0.380 & \textbf{0.470} & \textbf{0.430} & 0.407 & 0.370 \\
& & $8\times$ & 0.368 & 0.330 & 0.365 & 0.340 & \textbf{0.407} & \textbf{0.380} & 0.366 & 0.340 \\
& & $16\times$ & 0.289 & 0.250 & 0.297 & 0.260 & \textbf{0.381} & \textbf{0.330} & 0.358 & 0.320 \\
\addlinespace[2pt]
2Wiki & \textsc{True-20} & $4\times$ & \textbf{0.507} & \textbf{0.460} & 0.443 & 0.400 & 0.484 & 0.440 & 0.483 & 0.440 \\
& & $8\times$ & 0.360 & 0.310 & 0.382 & 0.340 & \textbf{0.478} & \textbf{0.440} & 0.439 & 0.410 \\
& & $16\times$ & 0.352 & 0.310 & 0.345 & 0.310 & \textbf{0.373} & \textbf{0.320} & 0.364 & 0.300 \\
\bottomrule
\end{tabular}
\vspace{3pt}
\begin{minipage}{0.97\textwidth}
\footnotesize
\textbf{Supporting-fact retention.}
Changing the compression order from Original to Reverse improves SF-R in all 12 dataset--setting--budget combinations, with an average gain of $+0.096$.
Because the retained content is fixed before downstream presentation, SF-R is identical under Canonical and Reverse presentation for a given compression order.
\end{minipage}
\end{table*}

Across the 12 settings, changing only the compression order from Original to Reverse while retaining canonical downstream presentation yields substantial improvements in downstream utility.
In contrast, changing only the downstream presentation order while keeping the Original-compressed content fixed produces considerably smaller and less consistent changes.
This decomposition supports the interpretation that context ordering primarily matters by changing which information survives sequential compression, rather than solely by repositioning already-retained evidence for the downstream model.
We use this experiment as a mechanism diagnostic for Reverse ordering rather than as a decomposition of the full \texttt{MORSE} search procedure.

\subsection{Detailed Baseline Definitions}
\label{app:baselines}

We provide detailed definitions of the baselines used throughout our experiments.
Unless otherwise stated, all sequential baselines use the same compressor, preprocessing pipeline, compression budget, and likelihood model.
Static ordering baselines differ only in the context order supplied to the compressor, while \textsc{RandomSearch-5} additionally performs compute-matched selection among multiple random orderings using the same objective $J_B$ as \texttt{MORSE}.
This isolates ordering and search effects from changes in the underlying compression procedure.

\paragraph{Independent.}
This baseline removes cross-context sequential conditioning.
Each compression unit is scored independently using query-conditioned information, without conditioning on units from previously processed contexts.
The resulting scores are therefore invariant to the permutation of the retrieved context collection.
This baseline represents a non-sequential compression strategy and distinguishes order-sensitive effects from those obtainable through independent relevance estimation alone.

\paragraph{Original.}
We preserve the context order provided by the dataset or retrieval pipeline and apply the sequential compressor directly.
We perform no additional relevance scoring or reranking.
This baseline measures sequential compression under the naturally supplied context order.

\paragraph{Random-5.}
We randomly permute the retrieved contexts before applying the same sequential compressor.
To reduce variance from any single random ordering, we evaluate five fixed random permutations and report the mean performance across them.
The same predetermined random permutations are used consistently across compression budgets for the corresponding examples.
This baseline estimates the performance of uninformed context ordering while preserving the same compression procedure.

\paragraph{RandomSearch-5.}
We sample five unique global random permutations of the retrieved context collection and apply the same sequential compressor independently to each candidate under the target budget.
Each compressed output is canonicalized in the same way as in \texttt{MORSE}, scored using the same compression-aware objective $J_B$, and the highest-scoring candidate is selected.
Unlike \textsc{Random-5}, which reports the mean over five random orderings, \textsc{RandomSearch-5} performs test-time selection and is therefore compute-matched to \texttt{MORSE} at $K=5$.
The two methods share the random candidates used for the matched comparison; \texttt{MORSE} replaces one random candidate with the Reverse anchor.

\paragraph{Length.}
We order contexts by increasing context length and then apply sequential compression.
This provides a simple query-independent ordering heuristic and tests whether the observed effects can be explained by a preference for shorter contexts appearing earlier in the compression history.

\paragraph{Forward.}
We rank contexts according to their forward conditional likelihood,
\begin{equation}
f(c_i;q)=\log p_\theta(c_i\mid q),
\end{equation}
and process contexts in decreasing order of this score.
Forward likelihood depends both on the relationship between the query and the context and on the intrinsic likelihood of the context itself.
It therefore provides a likelihood-based ordering baseline distinct from the reverse query-evidence criterion used by \texttt{MORSE}.

\paragraph{Reverse.}
We rank contexts independently according to reverse query likelihood,
\begin{equation}
r(c_i;q)=L_\theta(q)-L_\theta(q\mid c_i),
\end{equation}
which induces the same ordering as $\log p_\theta(q\mid c_i)$ for a fixed query.
We then process contexts from highest to lowest reverse score using the same sequential compressor.
This baseline corresponds to the evidence-first anchor used by \texttt{MORSE}, but performs no subsequent compression-aware search.
Comparing \textsc{Reverse} with \texttt{MORSE} therefore isolates the additional benefit of evaluating alternative candidate permutations through their compressed outputs.

\subsection{Detailed Results Corresponding to Table~\ref{tab:main_full_results}}
\label{app:table1_detailed}

Tables~\ref{tab:detail_hotpot_native_1p}-\ref{tab:detail_2wiki_true20_isd} report the three-decimal results underlying the rounded values in Table~\ref{tab:main_full_results}.
We separate each dataset, context setting, and compression procedure into its own table so that every table contains only one context setting and remains within the width of the two-column template.
Across the eight tables, \texttt{MORSE} achieves the highest SF-R in all 24 dataset--setting--procedure--ratio combinations.
Relative to the compute-matched \textsc{RandomSearch-5} baseline, \texttt{MORSE} improves SF-R in all 12 \textsc{1P} comparisons and all 12 \textsc{ISD} comparisons, showing that the Reverse anchor consistently improves evidence retention at a fixed five-candidate search budget.
The downstream pattern is more heterogeneous, especially under \textsc{ISD}: F1 and EM often improve, but the strongest static baseline can remain competitive in individual cells.
This distinction is consistent with treating SF-R as the direct measure of evidence preservation and F1/EM as downstream utility metrics that need not vary monotonically with retained annotated evidence.

\begin{table*}[t]
\centering
\caption{
\textbf{Detailed three-decimal results for HotpotQA, Native, under \textsc{1P}.}
SF-R, Answer F1, and EM are reported under $4\times$, $8\times$, and $16\times$ compression.
\texttt{MORSE} uses the final $K=5$ search budget, while \textsc{RandomSearch-5} is the compute-matched five-random-candidate search baseline.
The best result for each metric is shown in \textbf{bold} and the runner-up is \underline{underlined}.
Ranks use the underlying aggregate values before rounding to three decimals, with deterministic table-order tie-breaking for exact unrounded ties.
}
\label{tab:detail_hotpot_native_1p}
\fontsize{7.2}{8.2}\selectfont
\setlength{\tabcolsep}{3.4pt}
\renewcommand{\arraystretch}{1.08}
\begin{tabular}{@{}lccc ccc ccc@{}}
\toprule
\multirow{2}{*}{\textbf{Method}}
& \multicolumn{3}{c}{$4\times$}
& \multicolumn{3}{c}{$8\times$}
& \multicolumn{3}{c}{$16\times$} \\
\cmidrule(lr){2-4}
\cmidrule(lr){5-7}
\cmidrule(l){8-10}
& SF-R & F1 & EM
& SF-R & F1 & EM
& SF-R & F1 & EM \\
\midrule
Independent & 0.539 & 0.539 & 0.430 & 0.365 & 0.476 & 0.384 & 0.236 & 0.382 & 0.304 \\
Original & 0.588 & 0.551 & 0.434 & 0.436 & 0.505 & 0.394 & 0.290 & 0.414 & 0.318 \\
Random-5 & 0.609 & 0.559 & 0.442 & 0.451 & 0.493 & 0.384 & 0.301 & 0.414 & 0.327 \\
Length & 0.658 & 0.596 & 0.466 & 0.514 & 0.541 & 0.440 & 0.351 & 0.451 & 0.352 \\
Forward & 0.661 & 0.617 & 0.490 & 0.525 & 0.554 & \underline{0.456} & 0.369 & 0.467 & 0.368 \\
Reverse & 0.714 & 0.618 & 0.492 & 0.580 & 0.562 & 0.436 & 0.408 & 0.479 & 0.370 \\
\addlinespace[1pt]
\textsc{RandomSearch-5} & \underline{0.717} & \underline{0.622} & \textbf{0.501} & \underline{0.584} & \underline{0.571} & 0.454 & \underline{0.410} & \underline{0.481} & \underline{0.376} \\
\textbf{\texttt{MORSE}} & \textbf{0.737} & \textbf{0.624} & \underline{0.501} & \textbf{0.622} & \textbf{0.587} & \textbf{0.460} & \textbf{0.444} & \textbf{0.504} & \textbf{0.394} \\
\bottomrule
\end{tabular}
\end{table*}

\begin{table*}[t]
\centering
\caption{
\textbf{Detailed three-decimal results for HotpotQA, \textsc{True-20}, under \textsc{1P}.}
SF-R, Answer F1, and EM are reported under $4\times$, $8\times$, and $16\times$ compression.
\texttt{MORSE} uses the final $K=5$ search budget, while \textsc{RandomSearch-5} is the compute-matched five-random-candidate search baseline.
The best result for each metric is shown in \textbf{bold} and the runner-up is \underline{underlined}.
Ranks use the underlying aggregate values before rounding to three decimals, with deterministic table-order tie breaking for exact unrounded ties.
}
\label{tab:detail_hotpot_true20_1p}
\fontsize{7.2}{8.2}\selectfont
\setlength{\tabcolsep}{3.4pt}
\renewcommand{\arraystretch}{1.08}
\begin{tabular}{@{}lccc ccc ccc@{}}
\toprule
\multirow{2}{*}{\textbf{Method}}
& \multicolumn{3}{c}{$4\times$}
& \multicolumn{3}{c}{$8\times$}
& \multicolumn{3}{c}{$16\times$} \\
\cmidrule(lr){2-4}
\cmidrule(lr){5-7}
\cmidrule(l){8-10}
& SF-R & F1 & EM
& SF-R & F1 & EM
& SF-R & F1 & EM \\
\midrule
Independent & 0.683 & 0.644 & 0.540 & 0.472 & 0.501 & 0.450 & 0.284 & 0.410 & 0.360 \\
Original & 0.579 & 0.565 & 0.470 & 0.498 & 0.502 & 0.420 & 0.393 & 0.505 & 0.420 \\
Random-5 & 0.674 & 0.613 & 0.506 & 0.528 & 0.573 & 0.486 & 0.402 & 0.492 & 0.408 \\
Length & 0.660 & 0.617 & 0.520 & 0.558 & \underline{0.602} & 0.510 & 0.462 & \textbf{0.583} & \textbf{0.500} \\
Forward & 0.676 & 0.645 & \underline{0.550} & 0.585 & 0.581 & \textbf{0.520} & 0.475 & 0.538 & 0.470 \\
Reverse & 0.657 & 0.597 & 0.500 & 0.574 & 0.590 & 0.500 & 0.509 & 0.563 & 0.470 \\
\addlinespace[1pt]
\textsc{RandomSearch-5} & \underline{0.762} & \underline{0.656} & 0.550 & \underline{0.655} & 0.583 & 0.486 & \underline{0.520} & 0.560 & 0.464 \\
\textbf{\texttt{MORSE}} & \textbf{0.778} & \textbf{0.668} & \textbf{0.572} & \textbf{0.683} & \textbf{0.608} & \underline{0.516} & \textbf{0.576} & \underline{0.580} & \underline{0.484} \\
\bottomrule
\end{tabular}
\end{table*}

\begin{table*}[t]
\centering
\caption{
\textbf{Detailed three-decimal results for HotpotQA, Native, under \textsc{ISD}.}
SF-R, Answer F1, and EM are reported under $4\times$, $8\times$, and $16\times$ compression.
\texttt{MORSE} uses the final $K=5$ search budget, while \textsc{RandomSearch-5} is the compute-matched five-random-candidate search baseline.
The best result for each metric is shown in \textbf{bold} and the runner-up is \underline{underlined}.
Ranks use the underlying aggregate values before rounding to three decimals, with deterministic table-order tie breaking for exact unrounded ties.
}
\label{tab:detail_hotpot_native_isd}
\fontsize{7.2}{8.2}\selectfont
\setlength{\tabcolsep}{3.4pt}
\renewcommand{\arraystretch}{1.08}
\begin{tabular}{@{}lccc ccc ccc@{}}
\toprule
\multirow{2}{*}{\textbf{Method}}
& \multicolumn{3}{c}{$4\times$}
& \multicolumn{3}{c}{$8\times$}
& \multicolumn{3}{c}{$16\times$} \\
\cmidrule(lr){2-4}
\cmidrule(lr){5-7}
\cmidrule(l){8-10}
& SF-R & F1 & EM
& SF-R & F1 & EM
& SF-R & F1 & EM \\
\midrule
Independent & 0.532 & 0.605 & 0.520 & 0.307 & 0.442 & 0.380 & 0.197 & 0.414 & 0.340 \\
Original & 0.556 & 0.592 & 0.510 & 0.383 & 0.505 & 0.430 & 0.213 & 0.412 & 0.340 \\
Random-5 & 0.568 & 0.563 & 0.470 & 0.384 & 0.491 & 0.424 & 0.233 & 0.422 & 0.356 \\
Length & 0.707 & \underline{0.642} & \underline{0.540} & 0.494 & 0.579 & \underline{0.520} & 0.286 & 0.510 & \textbf{0.440} \\
Forward & \underline{0.708} & \textbf{0.679} & \textbf{0.610} & \underline{0.543} & \textbf{0.633} & \textbf{0.550} & 0.318 & 0.516 & 0.430 \\
Reverse & 0.699 & 0.633 & 0.540 & 0.543 & 0.579 & 0.490 & 0.336 & 0.522 & 0.420 \\
\addlinespace[1pt]
\textsc{RandomSearch-5} & 0.693 & 0.623 & 0.524 & 0.516 & 0.583 & 0.486 & \underline{0.346} & \underline{0.523} & \underline{0.440} \\
\textbf{\texttt{MORSE}} & \textbf{0.724} & 0.640 & 0.536 & \textbf{0.574} & \underline{0.616} & 0.510 & \textbf{0.374} & \textbf{0.532} & 0.436 \\
\bottomrule
\end{tabular}
\end{table*}

\begin{table*}[t]
\centering
\caption{
\textbf{Detailed three-decimal results for HotpotQA, \textsc{True-20}, under \textsc{ISD}.}
SF-R, Answer F1, and EM are reported under $4\times$, $8\times$, and $16\times$ compression.
\texttt{MORSE} uses the final $K=5$ search budget, while \textsc{RandomSearch-5} is the compute-matched five-random-candidate search baseline.
The best result for each metric is shown in \textbf{bold} and the runner-up is \underline{underlined}.
Ranks use the underlying aggregate values before rounding to three decimals, with deterministic table-order tie breaking for exact unrounded ties.
}
\label{tab:detail_hotpot_true20_isd}
\fontsize{7.2}{8.2}\selectfont
\setlength{\tabcolsep}{3.4pt}
\renewcommand{\arraystretch}{1.08}
\begin{tabular}{@{}lccc ccc ccc@{}}
\toprule
\multirow{2}{*}{\textbf{Method}}
& \multicolumn{3}{c}{$4\times$}
& \multicolumn{3}{c}{$8\times$}
& \multicolumn{3}{c}{$16\times$} \\
\cmidrule(lr){2-4}
\cmidrule(lr){5-7}
\cmidrule(l){8-10}
& SF-R & F1 & EM
& SF-R & F1 & EM
& SF-R & F1 & EM \\
\midrule
Independent & 0.683 & 0.649 & 0.540 & 0.472 & 0.482 & 0.420 & 0.284 & 0.439 & 0.380 \\
Original & 0.663 & 0.645 & 0.530 & 0.546 & 0.591 & 0.510 & 0.392 & 0.488 & 0.410 \\
Random-5 & 0.699 & 0.640 & 0.540 & 0.525 & 0.578 & 0.500 & 0.387 & 0.507 & 0.424 \\
Length & 0.748 & \textbf{0.691} & \textbf{0.600} & 0.610 & \underline{0.628} & \underline{0.550} & 0.490 & 0.571 & \underline{0.510} \\
Forward & 0.778 & \underline{0.689} & \underline{0.590} & \underline{0.661} & \textbf{0.656} & \textbf{0.600} & 0.527 & \textbf{0.612} & \textbf{0.550} \\
Reverse & \underline{0.784} & 0.666 & 0.570 & 0.653 & 0.589 & 0.510 & \underline{0.547} & 0.560 & 0.490 \\
\addlinespace[1pt]
\textsc{RandomSearch-5} & 0.772 & 0.670 & 0.566 & 0.627 & 0.625 & 0.522 & 0.505 & 0.573 & 0.486 \\
\textbf{\texttt{MORSE}} & \textbf{0.797} & 0.683 & 0.580 & \textbf{0.676} & 0.616 & 0.510 & \textbf{0.567} & \underline{0.586} & 0.484 \\
\bottomrule
\end{tabular}
\end{table*}

\begin{table*}[t]
\centering
\caption{
\textbf{Detailed three-decimal results for 2WikiMultiHopQA, Native, under \textsc{1P}.}
SF-R, Answer F1, and EM are reported under $4\times$, $8\times$, and $16\times$ compression.
\texttt{MORSE} uses the final $K=5$ search budget, while \textsc{RandomSearch-5} is the compute-matched five-random-candidate search baseline.
The best result for each metric is shown in \textbf{bold} and the runner-up is \underline{underlined}.
Ranks use the underlying aggregate values before rounding to three decimals, with deterministic table-order tie breaking for exact unrounded ties.
}
\label{tab:detail_2wiki_native_1p}
\fontsize{7.2}{8.2}\selectfont
\setlength{\tabcolsep}{3.4pt}
\renewcommand{\arraystretch}{1.08}
\begin{tabular}{@{}lccc ccc ccc@{}}
\toprule
\multirow{2}{*}{\textbf{Method}}
& \multicolumn{3}{c}{$4\times$}
& \multicolumn{3}{c}{$8\times$}
& \multicolumn{3}{c}{$16\times$} \\
\cmidrule(lr){2-4}
\cmidrule(lr){5-7}
\cmidrule(l){8-10}
& SF-R & F1 & EM
& SF-R & F1 & EM
& SF-R & F1 & EM \\
\midrule
Independent & 0.364 & 0.450 & 0.420 & 0.163 & 0.356 & 0.340 & 0.083 & 0.327 & 0.310 \\
Original & 0.456 & 0.438 & 0.420 & 0.255 & 0.368 & 0.330 & 0.130 & 0.289 & 0.250 \\
Random-5 & 0.453 & 0.433 & 0.388 & 0.243 & 0.368 & 0.334 & 0.135 & 0.319 & 0.274 \\
Length & 0.397 & 0.404 & 0.360 & 0.217 & 0.399 & 0.340 & 0.115 & 0.303 & 0.250 \\
Forward & 0.422 & 0.433 & 0.390 & 0.217 & 0.335 & 0.290 & 0.122 & 0.352 & 0.290 \\
Reverse & 0.546 & 0.470 & 0.430 & \underline{0.354} & \underline{0.407} & \underline{0.380} & 0.214 & \underline{0.381} & 0.330 \\
\addlinespace[1pt]
\textsc{RandomSearch-5} & \underline{0.550} & \underline{0.476} & \underline{0.434} & 0.349 & 0.398 & 0.360 & \underline{0.226} & 0.380 & \underline{0.342} \\
\textbf{\texttt{MORSE}} & \textbf{0.588} & \textbf{0.486} & \textbf{0.444} & \textbf{0.375} & \textbf{0.422} & \textbf{0.382} & \textbf{0.240} & \textbf{0.397} & \textbf{0.354} \\
\bottomrule
\end{tabular}
\end{table*}

\begin{table*}[t]
\centering
\caption{
\textbf{Detailed three-decimal results for 2WikiMultiHopQA, \textsc{True-20}, under \textsc{1P}.}
SF-R, Answer F1, and EM are reported under $4\times$, $8\times$, and $16\times$ compression.
\texttt{MORSE} uses the final $K=5$ search budget, while \textsc{RandomSearch-5} is the compute-matched five-random-candidate search baseline.
The best result for each metric is shown in \textbf{bold} and the runner-up is \underline{underlined}.
Ranks use the underlying aggregate values before rounding to three decimals, with deterministic table-order tie breaking for exact unrounded ties.
}
\label{tab:detail_2wiki_true20_1p}
\fontsize{7.2}{8.2}\selectfont
\setlength{\tabcolsep}{3.4pt}
\renewcommand{\arraystretch}{1.08}
\begin{tabular}{@{}lccc ccc ccc@{}}
\toprule
\multirow{2}{*}{\textbf{Method}}
& \multicolumn{3}{c}{$4\times$}
& \multicolumn{3}{c}{$8\times$}
& \multicolumn{3}{c}{$16\times$} \\
\cmidrule(lr){2-4}
\cmidrule(lr){5-7}
\cmidrule(l){8-10}
& SF-R & F1 & EM
& SF-R & F1 & EM
& SF-R & F1 & EM \\
\midrule
Independent & 0.481 & 0.452 & 0.430 & 0.272 & 0.379 & 0.370 & 0.120 & 0.292 & 0.270 \\
Original & 0.533 & 0.507 & \underline{0.460} & 0.397 & 0.360 & 0.310 & 0.225 & 0.352 & 0.310 \\
Random-5 & 0.576 & 0.474 & 0.418 & 0.407 & 0.415 & 0.366 & 0.227 & 0.342 & 0.304 \\
Length & 0.609 & \underline{0.517} & 0.460 & 0.352 & 0.411 & 0.350 & 0.195 & 0.348 & 0.310 \\
Forward & 0.604 & 0.457 & 0.410 & 0.382 & 0.421 & 0.360 & 0.212 & 0.325 & 0.260 \\
Reverse & 0.591 & 0.484 & 0.440 & 0.466 & 0.478 & \underline{0.440} & 0.317 & 0.373 & 0.320 \\
\addlinespace[1pt]
\textsc{RandomSearch-5} & \underline{0.637} & \textbf{0.521} & \textbf{0.466} & \underline{0.475} & \underline{0.481} & 0.440 & \underline{0.325} & \underline{0.417} & \textbf{0.376} \\
\textbf{\texttt{MORSE}} & \textbf{0.657} & 0.507 & 0.452 & \textbf{0.512} & \textbf{0.506} & \textbf{0.466} & \textbf{0.354} & \textbf{0.422} & \underline{0.368} \\
\bottomrule
\end{tabular}
\end{table*}

\begin{table*}[t]
\centering
\caption{
\textbf{Detailed three-decimal results for 2WikiMultiHopQA, Native, under \textsc{ISD}.}
SF-R, Answer F1, and EM are reported under $4\times$, $8\times$, and $16\times$ compression.
\texttt{MORSE} uses the final $K=5$ search budget, while \textsc{RandomSearch-5} is the compute-matched five-random-candidate search baseline.
The best result for each metric is shown in \textbf{bold} and the runner-up is \underline{underlined}.
Ranks use the underlying aggregate values before rounding to three decimals, with deterministic table-order tie breaking for exact unrounded ties.
}
\label{tab:detail_2wiki_native_isd}
\fontsize{7.2}{8.2}\selectfont
\setlength{\tabcolsep}{3.4pt}
\renewcommand{\arraystretch}{1.08}
\begin{tabular}{@{}lccc ccc ccc@{}}
\toprule
\multirow{2}{*}{\textbf{Method}}
& \multicolumn{3}{c}{$4\times$}
& \multicolumn{3}{c}{$8\times$}
& \multicolumn{3}{c}{$16\times$} \\
\cmidrule(lr){2-4}
\cmidrule(lr){5-7}
\cmidrule(l){8-10}
& SF-R & F1 & EM
& SF-R & F1 & EM
& SF-R & F1 & EM \\
\midrule
Independent & 0.364 & \textbf{0.473} & \textbf{0.420} & 0.163 & 0.346 & 0.320 & 0.083 & 0.306 & 0.290 \\
Original & 0.407 & 0.441 & 0.390 & 0.217 & 0.372 & 0.330 & 0.092 & 0.349 & 0.310 \\
Random-5 & 0.405 & 0.425 & 0.374 & 0.217 & 0.374 & 0.330 & 0.083 & 0.305 & 0.264 \\
Length & 0.362 & 0.376 & 0.350 & 0.188 & 0.392 & 0.340 & 0.070 & 0.326 & 0.270 \\
Forward & 0.392 & 0.349 & 0.320 & 0.177 & 0.365 & 0.310 & 0.062 & 0.329 & 0.270 \\
Reverse & \underline{0.551} & 0.391 & 0.350 & \underline{0.309} & 0.369 & 0.320 & 0.162 & \underline{0.374} & \underline{0.330} \\
\addlinespace[1pt]
\textsc{RandomSearch-5} & 0.516 & \underline{0.449} & \underline{0.396} & 0.292 & \underline{0.403} & \underline{0.354} & \underline{0.170} & 0.367 & 0.330 \\
\textbf{\texttt{MORSE}} & \textbf{0.560} & 0.436 & 0.388 & \textbf{0.336} & \textbf{0.411} & \textbf{0.356} & \textbf{0.185} & \textbf{0.393} & \textbf{0.352} \\
\bottomrule
\end{tabular}
\end{table*}

\begin{table*}[t]
\centering
\caption{
\textbf{Detailed three-decimal results for 2WikiMultiHopQA, \textsc{True-20}, under \textsc{ISD}.}
SF-R, Answer F1, and EM are reported under $4\times$, $8\times$, and $16\times$ compression.
\texttt{MORSE} uses the final $K=5$ search budget, while \textsc{RandomSearch-5} is the compute-matched five-random-candidate search baseline.
The best result for each metric is shown in \textbf{bold} and the runner-up is \underline{underlined}.
Ranks use the underlying aggregate values before rounding to three decimals, with deterministic table-order tie breaking for exact unrounded ties.
}
\label{tab:detail_2wiki_true20_isd}
\fontsize{7.2}{8.2}\selectfont
\setlength{\tabcolsep}{3.4pt}
\renewcommand{\arraystretch}{1.08}
\begin{tabular}{@{}lccc ccc ccc@{}}
\toprule
\multirow{2}{*}{\textbf{Method}}
& \multicolumn{3}{c}{$4\times$}
& \multicolumn{3}{c}{$8\times$}
& \multicolumn{3}{c}{$16\times$} \\
\cmidrule(lr){2-4}
\cmidrule(lr){5-7}
\cmidrule(l){8-10}
& SF-R & F1 & EM
& SF-R & F1 & EM
& SF-R & F1 & EM \\
\midrule
Independent & 0.481 & 0.466 & 0.420 & 0.272 & 0.399 & 0.370 & 0.120 & 0.325 & 0.300 \\
Original & 0.589 & 0.475 & 0.410 & 0.377 & 0.413 & 0.350 & 0.202 & 0.365 & 0.320 \\
Random-5 & 0.549 & 0.455 & 0.392 & 0.334 & 0.389 & 0.348 & 0.151 & 0.315 & 0.282 \\
Length & 0.520 & 0.461 & 0.400 & 0.335 & 0.370 & 0.320 & 0.135 & 0.346 & 0.300 \\
Forward & 0.525 & 0.443 & 0.370 & 0.318 & 0.379 & 0.350 & 0.133 & 0.310 & 0.260 \\
Reverse & \underline{0.641} & \textbf{0.495} & \textbf{0.430} & \underline{0.501} & 0.416 & 0.370 & 0.255 & 0.346 & 0.310 \\
\addlinespace[1pt]
\textsc{RandomSearch-5} & 0.624 & 0.477 & 0.426 & 0.436 & \underline{0.432} & \underline{0.388} & \underline{0.261} & \underline{0.388} & \underline{0.358} \\
\textbf{\texttt{MORSE}} & \textbf{0.642} & \underline{0.482} & \underline{0.428} & \textbf{0.516} & \textbf{0.454} & \textbf{0.404} & \textbf{0.307} & \textbf{0.404} & \textbf{0.360} \\
\bottomrule
\end{tabular}
\end{table*}

The detailed results reinforce two complementary conclusions.
First, the evidence-retention advantage of \texttt{MORSE} is unusually stable: the method is first-ranked in SF-R for every setting shown above, including both datasets, both context regimes, both compression procedures, and all three budgets.
Second, downstream answer quality is less uniform, particularly for \textsc{ISD}, where a method can preserve more annotated supporting facts without necessarily maximizing F1 or EM in the same cell.
We therefore use the detailed appendix tables to support the evidence-preservation claim directly while keeping downstream conclusions appropriately calibrated.

\clearpage

\subsection{Preemption Correction Analysis}
\label{app:preemption_correction}

We further examine whether the ordering methods themselves correct the harmful preemptor--evidence relations identified by the controlled pair-swap experiment.
We use exactly the same 399 measured pairs $(c_p,c_e)$ from Section \ref{sec:pair_swap}, where $c_p$ denotes the selected strongest non-gold preemptor and $c_e$ denotes the corresponding gold evidence carrier.
Of these pairs, 183 initially satisfy $c_p \prec c_e$ in the original context order, while the remaining 216 already satisfy the favorable relation $c_e \prec c_p$.

For the initially harmful subset
\[
\mathcal{B}
=
\{(c_p,c_e): c_p \prec_{\mathrm{Original}} c_e\},
\]
we define the preemption correction rate of method $M$ as
\[
\mathrm{PCR}(M)
=
\frac{1}{|\mathcal{B}|}
\sum_{(c_p,c_e)\in\mathcal{B}}
\mathbf{1}
\left[
c_e \prec_M c_p
\right].
\]
For the initially favorable subset, we analogously report the fraction for which the method preserves $c_e \prec c_p$.
We additionally report the final harmful-order rate over all 399 measured pairs.
For \textsc{MORSE-5} and \textsc{RandomSearch-5}, all quantities are computed separately for seeds 42--46 and summarized by the mean and sample standard deviation.

\begin{table}[t]
\centering
\caption{
\textbf{Ratio-specific preemption correction rates.}
Correction is the fraction of the 183 initially harmful pairs for which the selected compression order changes $c_p \prec c_e$ to $c_e \prec c_p$.
Values for \textsc{MORSE-5} and \textsc{RandomSearch-5} are mean $\pm$ sample standard deviation over five seeds.
}
\label{tab:preemption_correction_full}
\small
\setlength{\tabcolsep}{6.0pt}
\renewcommand{\arraystretch}{1.08}
\begin{tabular}{@{}lcccc@{}}
\toprule
\textbf{Method}
& \textbf{$4\times$}
& \textbf{$8\times$}
& \textbf{$16\times$}
& \textbf{Mean} \\
\midrule
Original
& 0.000
& 0.000
& 0.000
& 0.000 \\
Reverse
& 0.869
& 0.869
& 0.869
& 0.869 \\
\textsc{RandomSearch-5}
& $0.554{\pm}0.031$
& $0.550{\pm}0.036$
& $0.558{\pm}0.016$
& $0.554{\pm}0.021$ \\
\textsc{MORSE-5}
& $0.646{\pm}0.015$
& $0.645{\pm}0.049$
& $0.678{\pm}0.024$
& $0.656{\pm}0.020$ \\
\bottomrule
\end{tabular}
\end{table}

Across all three compression ratios, \textsc{MORSE-5} corrects harmful relations more frequently than the compute-matched \textsc{RandomSearch-5}.
The paired differences are $+9.2$ points at $4\times$, $+9.5$ points at $8\times$, and $+11.9$ points at $16\times$ compression.
The corresponding pair-bootstrap 95\% confidence intervals are $[5.9,12.5]$, $[5.9,13.2]$, and $[8.3,15.6]$ percentage points, respectively.
Averaged across ratios, the correction advantage is $+10.2$ points with a 95\% confidence interval of $[7.5,12.9]$ points.
Reverse yields the highest correction rate at $86.9\%$, consistent with its explicit evidence-first construction.
MORSE therefore preserves a substantial part of this pairwise evidence-first bias while allowing compression-aware $J_B$ selection to choose alternative global permutations.

\begin{table}[t]
\centering
\caption{
\textbf{Preservation and final prevalence of harmful preemptor--evidence relations.}
Good-order preservation is measured over the 216 pairs already satisfying $c_e \prec c_p$ in the original order.
Harmful-order rate is the fraction of all 399 measured pairs whose final compression order satisfies $c_p \prec c_e$.
}
\label{tab:preemption_preservation}
\small
\setlength{\tabcolsep}{7.0pt}
\renewcommand{\arraystretch}{1.08}
\begin{tabular}{@{}lcc@{}}
\toprule
\textbf{Method}
& \textbf{Good-order preservation $\uparrow$}
& \textbf{Harmful-order rate $\downarrow$} \\
\midrule
Original
& 1.000
& 0.459 \\
Reverse
& 0.866
& 0.133 \\
\textsc{RandomSearch-5}
& $0.520{\pm}0.025$
& $0.464{\pm}0.021$ \\
\textsc{MORSE-5}
& $0.630{\pm}0.016$
& $0.358{\pm}0.012$ \\
\bottomrule
\end{tabular}
\end{table}

The correction advantage of MORSE is not explained by indiscriminate permutation.
Among pairs that already have the favorable evidence-first relation, \textsc{MORSE-5} preserves that relation in $63.0\%$ of cases, compared with $52.0\%$ for \textsc{RandomSearch-5}.
Consequently, MORSE reduces the final harmful-order rate by $10.1$ percentage points relative to the original order, from $45.9\%$ to $35.8\%$.
In contrast, the harmful-order rate under \textsc{RandomSearch-5} is $46.4\%$, essentially unchanged from the original ordering.
We finally connect these naturally occurring MORSE corrections back to the controlled intervention in Section \ref{sec:pair_swap}.
Using a strict majority over the 15 seed--ratio decisions, MORSE usually corrects 140 of the 183 initially harmful pairs and usually leaves 43 uncorrected.
Pairs that MORSE usually corrects have a pair-swap rescue rate of $45.2\%$, compared with $21.7\%$ among pairs it usually leaves uncorrected.
Their mean measured suppression is also somewhat larger, $8.08$ versus $7.63$, and the mean pair-swap intervention effect is substantially larger, with $\Delta$SF-R of $+20.6$ versus $+4.9$ percentage points.
Thus, MORSE corrections are preferentially associated with pairs for which the controlled evidence-first intervention has a larger consequence for evidence survival.
This comparison is descriptive rather than causal with respect to MORSE's own correction decisions, since the causal intervention is the controlled pair swap itself.
Taken together, these results distinguish the roles of the two components of MORSE.
The Reverse anchor most directly enforces favorable local preemptor--evidence relations, whereas compression-aware selection optimizes the evidence retained by the complete compressed context set.
MORSE therefore corrects substantially more harmful relations than matched random search while retaining the flexibility to sacrifice individual pairwise relations when another permutation obtains a better global compression objective.

\subsection{Selection-Objective Ablation}
\label{app:jb_ablation}

We provide the complete analysis of the compression-aware selection objective used in Section~\ref{sec:jb_ablation}.
The experiment reuses the existing candidate trajectories from all 12 Qwen2.5-0.5B 1P settings, covering HotpotQA and 2WikiMultiHopQA under Native and TRUE-20 contexts at $4\times$, $8\times$, and $16\times$ compression.
No candidate permutation is regenerated and no compression is rerun.
Thus, the compressed outputs are identical across all selector comparisons.

\paragraph{Selection objectives.}
For each candidate permutation $\pi$, let
$\widehat{\mathcal C}_{\pi,B}$ denote its compressed output restored to canonical presentation order.
The MORSE objective is
\begin{equation}
    J_B(\pi)
    =
    L_\theta(q)
    -
    L_\theta(q\mid\widehat{\mathcal C}_{\pi,B}),
\end{equation}
which measures how strongly the retained content supports the query.
We compare it with the forward conditional likelihood
\begin{equation}
    J_B^{\mathrm{fwd}}(\pi)
    =
    \log p_\theta(
        \widehat{\mathcal C}_{\pi,B}\mid q
    ),
\end{equation}
which instead measures how likely the retained content is given the query.
Because summed forward log-likelihood may depend on output length, we additionally evaluate
\begin{equation}
    J_B^{\mathrm{fwd\text{-}norm}}(\pi)
    =
    \frac{
        \log p_\theta(
            \widehat{\mathcal C}_{\pi,B}\mid q
        )
    }{
        N_{\pi,B}
    },
\end{equation}
where $N_{\pi,B}$ is the number of scored target tokens.

\paragraph{Controlled candidate pools.}
We evaluate each selector on two fixed candidate pools.
The MORSE pool contains
$\{\text{Reverse},R_1,R_2,R_3,R_4\}$,
whereas the random pool contains
$\{R_1,R_2,R_3,R_4,R_5\}$.
All candidate outputs are held fixed when changing the selector.
The random pool therefore tests whether the advantage of $J_B$ persists even when no Reverse anchor is present.

\begin{table}[t]
\centering
\caption{
\textbf{Selection-objective ablation.}
Values are equal-setting averages over the 12 1P settings.
Oracle denotes selection using the true SF-R of each candidate and is used only as an analysis upper bound.
}
\label{tab:jb_selector_full}
\small
\setlength{\tabcolsep}{4.8pt}
\renewcommand{\arraystretch}{1.08}
\begin{tabular}{@{}lcccc@{}}
\toprule
\textbf{Selector}
& \textbf{MORSE pool}
& \textbf{Random pool}
& \textbf{MORSE regret}
& \textbf{Random regret} \\
\midrule
Forward
& 0.4898
& 0.4650
& 0.1599
& 0.1568 \\
Forward-Norm.
& 0.4927
& 0.4698
& 0.1570
& 0.1520 \\
Reverse $J_B$
& \textbf{0.5471}
& \textbf{0.5175}
& \textbf{0.1026}
& \textbf{0.1043} \\
Oracle
& 0.6497
& 0.6218
& --
& -- \\
\bottomrule
\end{tabular}
\end{table}

Using reverse $J_B$ instead of Forward increases selected SF-R by
$5.73$ points in the MORSE pool
(95\% CI $[4.78,6.69]$)
and by $5.26$ points in the random pool
(95\% CI $[4.42,6.09]$).
The improvements remain $5.45$ and $4.77$ points, respectively, after normalizing the forward score by output length.
Reverse $J_B$ also substantially reduces regret relative to the SF-R oracle.

\paragraph{Correlation with retained evidence.}
We next test whether the three objectives rank alternative compressed outputs according to their actual evidence retention.
For each example and compression ratio, we pool the existing candidates across all five search seeds, deduplicate identical canonical compressed outputs, and compute the rank correlation between each selector score and SF-R.
Correlations are computed \emph{within each example}, rather than by pooling raw likelihood values across queries, because likelihood scales are not directly comparable across examples.
Cases with no SF-R variation are excluded rather than assigned zero correlation.

\begin{table}[t]
\centering
\caption{
\textbf{Within-example correlation between candidate scores and SF-R.}
Values are equal-setting means over the 12 1P settings.
}
\label{tab:jb_correlation}
\small
\setlength{\tabcolsep}{7.0pt}
\renewcommand{\arraystretch}{1.08}
\begin{tabular}{@{}lcc@{}}
\toprule
\textbf{Score}
& \textbf{Spearman $\rho$}
& \textbf{Kendall $\tau_b$} \\
\midrule
Forward
& 0.230 & 0.191 \\
Forward-Norm.
& 0.251 & 0.209 \\
Reverse $J_B$
& \textbf{0.401} & \textbf{0.335} \\
\bottomrule
\end{tabular}
\end{table}

Reverse $J_B$ yields substantially stronger within-example association with SF-R.
Its Spearman correlation exceeds Forward by $0.171$
(95\% CI $[0.147,0.195]$)
and Forward-Norm. by $0.150$
(95\% CI $[0.127,0.173]$).
The same pattern holds for Kendall $\tau_b$, with corresponding improvements of $0.144$ and $0.127$.
Correlations are defined for 2,249 of the 2,400 example--ratio cases; undefined cases arise primarily when all available candidates obtain the same SF-R.

\paragraph{Random-only control.}
To ensure that the correlation advantage is not caused by including the Reverse anchor itself, we repeat the analysis using only random-permutation candidates.
The resulting Spearman correlations are $0.229$, $0.250$, and $0.397$ for Forward, Forward-Norm., and reverse $J_B$, respectively, with corresponding Kendall correlations of $0.191$, $0.208$, and $0.332$.
Thus, the stronger association of $J_B$ with retained evidence persists even when the Reverse candidate is excluded entirely.

\paragraph{Oracle agreement.}
As an additional diagnostic, we measure how often each selector chooses a candidate belonging to the set of candidates with maximal SF-R.
In the MORSE pool, oracle top-1 agreement is $66.2\%$ for Forward, $66.9\%$ for Forward-Norm., and $76.4\%$ for reverse $J_B$.
The corresponding values in the random pool are $66.5\%$, $67.5\%$, and $76.1\%$.
Together with the lower oracle regret and stronger rank correlations, these results show that the reverse retained-evidence objective provides a substantially better unsupervised proxy for SF-R than forward conditional likelihood.

\subsection{Cross-Model and Cross-Dataset Generalization}
\label{app:generalization}

We provide the complete cross-model and cross-dataset generalization experiment corresponding to Section~\ref{sec:generalization}.
The experiment follows a $2\times2$ design over two compression scorers and two multi-hop QA datasets.
The scorers are Qwen2.5-0.5B-Instruct and OLMo-2-0425-1B-Instruct.
The datasets are HotpotQA TRUE-20 and MuSiQue, with the context count fixed to 20 in every evaluated example.
Only the primary 1P compressor is used, with compression ratios of $4\times$, $8\times$, and $16\times$ and the default MORSE candidate budget $K=5$.
The downstream reader remains Qwen2.5-7B-Instruct in every condition.
Thus, the experiment changes only the compression scorer and dataset while preserving the remaining compression and downstream evaluation protocol.

\paragraph{Dataset construction.}
For HotpotQA, we reuse the exact fixed 100-example TRUE-20 population from the main experiments.
For MuSiQue, we use a prospectively sampled 100-example subset of the official answerable development split.
We restrict the eligible population to examples with exactly 20 native paragraph contexts that remain nonempty under the frozen preprocessing procedure, yielding 1,963 eligible examples before sampling.
The resulting fixed subset contains 53 two-hop, 29 three-hop, and 18 four-hop questions.
We constructed and froze the MuSiQue population before any scientific inference.

\paragraph{Evidence-retention metric.}
HotpotQA continues to use Supporting-Fact Recall (SF-R) based on its sentence-level supporting-fact annotations.
MuSiQue instead provides paragraph-level support annotations.
We therefore define Supporting-Paragraph Recall (SP-R) as
\begin{equation}
    \mathrm{SP\text{-}R}
    =
    \frac{
        \#\{c \in \mathcal{G}: |\widetilde{c}| > 0\}
    }{
        |\mathcal{G}|
    },
\end{equation}
where $\mathcal{G}$ is the set of annotated supporting paragraphs and $\widetilde{c}$ denotes the retained sentences originating from supporting paragraph $c$ after compression.
A supporting paragraph is therefore counted as preserved if at least one complete sentence from that paragraph survives.
We additionally report Supporting-Paragraph Token Retention (SP-TR), the fraction of token mass from gold supporting paragraphs that remains after compression, as a secondary diagnostic.
SF-R and SP-R are not pooled across datasets.

\paragraph{Scoring models.}
The original scorer is Qwen2.5-0.5B-Instruct at revision \texttt{7ae557604adf67be50417f59c2c2f167def9a775}.
The transferred scorer is OLMo-2-0425-1B-Instruct at revision \texttt{48d788eca847d4d7548f375ad03d3c9312f6139e}.
For each scorer, all likelihood-derived quantities use the same active model, including sequential chain-QMI scores, Forward and Reverse ordering scores, and the compression-aware objective $J_B$.
The raw contexts, sentence segmentation, and compression-budget accounting are held fixed across scorers.
The downstream reader is Qwen2.5-7B-Instruct at revision \texttt{a09a35458c702b33eeacc393d103063234e8bc28}.

\paragraph{Search protocol.}
MORSE-5 evaluates the Reverse anchor together with four random global permutations.
RandomSearch-5 evaluates five random global permutations under the same compression-aware objective.
The first four random candidates are shared exactly between the two methods.
Search results are averaged over deterministic seeds 42--46.
All retained content is restored to canonical presentation order before downstream QA.

\begin{table*}[t]
\centering
\caption{
\textbf{Complete generalization results for the three primary ordering methods.}
Retention denotes SF-R for HotpotQA and SP-R for MuSiQue.
MORSE-5 and RandomSearch-5 values are five-seed means.
The downstream reader is fixed to Qwen2.5-7B-Instruct in every setting.
}
\label{tab:generalization_full}
\small
\setlength{\tabcolsep}{2.5pt}
\renewcommand{\arraystretch}{1.08}
\begin{tabular}{@{}ll l ccc ccc ccc@{}}
\toprule
\textbf{Scorer}
& \textbf{Dataset}
& \textbf{Method}
& \multicolumn{3}{c}{$\mathbf{4\times}$}
& \multicolumn{3}{c}{$\mathbf{8\times}$}
& \multicolumn{3}{c}{$\mathbf{16\times}$} \\
\cmidrule(lr){4-6}
\cmidrule(lr){7-9}
\cmidrule(lr){10-12}
&
&
&
\textbf{Ret.}
& \textbf{F1}
& \textbf{EM}
& \textbf{Ret.}
& \textbf{F1}
& \textbf{EM}
& \textbf{Ret.}
& \textbf{F1}
& \textbf{EM} \\
\midrule

Qwen
& HotpotQA
& Reverse
& 0.6572 & 0.5965 & 0.500
& 0.5735 & 0.5900 & 0.500
& 0.5093 & 0.5628 & 0.470 \\

&
&
RandomSearch-5
& 0.7622 & 0.6558 & 0.550
& 0.6547 & 0.5833 & 0.486
& 0.5202 & 0.5600 & 0.464 \\

&
&
MORSE-5
& 0.7780 & 0.6682 & 0.572
& 0.6827 & 0.6082 & 0.516
& 0.5761 & 0.5804 & 0.484 \\

\midrule

OLMo
& HotpotQA
& Reverse
& 0.5515 & 0.5476 & 0.460
& 0.4582 & 0.4742 & 0.400
& 0.3773 & 0.4318 & 0.360 \\

&
&
RandomSearch-5
& 0.6089 & 0.5893 & 0.484
& 0.4850 & 0.5457 & 0.454
& 0.3783 & 0.4833 & 0.394 \\

&
&
MORSE-5
& 0.6432 & 0.6003 & 0.502
& 0.5281 & 0.5471 & 0.456
& 0.4322 & 0.5011 & 0.406 \\

\midrule

Qwen
& MuSiQue
& Reverse
& 0.7800 & 0.3053 & 0.190
& 0.6800 & 0.2955 & 0.200
& 0.5017 & 0.2990 & 0.200 \\

&
&
RandomSearch-5
& 0.8283 & 0.3223 & 0.222
& 0.6927 & 0.3066 & 0.208
& 0.5550 & 0.2698 & 0.178 \\

&
&
MORSE-5
& 0.8407 & 0.3250 & 0.220
& 0.7247 & 0.3177 & 0.226
& 0.5728 & 0.2789 & 0.188 \\

\midrule

OLMo
& MuSiQue
& Reverse
& 0.6492 & 0.3082 & 0.190
& 0.5083 & 0.2170 & 0.130
& 0.3683 & 0.1680 & 0.120 \\

&
&
RandomSearch-5
& 0.7802 & 0.3053 & 0.194
& 0.6028 & 0.2361 & 0.148
& 0.4723 & 0.2326 & 0.148 \\

&
&
MORSE-5
& 0.7698 & 0.3013 & 0.194
& 0.6207 & 0.2358 & 0.140
& 0.4865 & 0.2194 & 0.138 \\

\bottomrule
\end{tabular}
\end{table*}

\paragraph{Retention generalization.}
Table~\ref{tab:generalization_full} shows that MORSE improves evidence retention over Reverse at every ratio in all four scorer--dataset combinations.
This pattern remains strong in every transferred condition.
With OLMo on HotpotQA, MORSE improves over Reverse by $9.17$, $7.00$, and $5.48$ percentage points at $4\times$, $8\times$, and $16\times$ compression, respectively.
With Qwen on MuSiQue, the corresponding improvements are $6.07$, $4.47$, and $7.12$ points.
Under the joint OLMo--MuSiQue shift, the gains over Reverse remain $12.07$, $11.23$, and $11.82$ points.
Thus, the benefit of compression-aware exploration beyond the static Reverse ordering transfers strongly across both scorer and dataset changes.

\begin{table}[t]
\centering
\caption{
\textbf{Retention differences between MORSE-5 and RandomSearch-5 in the three transferred conditions.}
Intervals are paired example-level bootstrap 95\% confidence intervals.
The final column averages the three compression ratios equally.
}
\label{tab:generalization_randomsearch_delta}
\small
\setlength{\tabcolsep}{4.0pt}
\renewcommand{\arraystretch}{1.08}
\begin{tabular}{@{}llcccc@{}}
\toprule
\textbf{Scorer}
& \textbf{Dataset}
& $\mathbf{4\times}$
& $\mathbf{8\times}$
& $\mathbf{16\times}$
& \textbf{Mean} \\
\midrule
OLMo
& HotpotQA
& $+.0343$
& $+.0432$
& $+.0538$
& $+.0438$ \\
Qwen
& MuSiQue
& $+.0123$
& $+.0320$
& $+.0178$
& $+.0207$ \\
OLMo
& MuSiQue
& $-.0103$
& $+.0178$
& $+.0142$
& $+.0072$ \\
\bottomrule
\end{tabular}
\end{table}

Relative to the compute-matched RandomSearch-5 baseline, MORSE transfers most clearly under the individual scorer and dataset shifts.
With OLMo on HotpotQA, MORSE improves retention at all three ratios, with an equal-ratio mean difference of $+4.38$ points and a 95\% confidence interval of $[+2.34,+6.59]$ points.
With Qwen on MuSiQue, the equal-ratio advantage is $+2.07$ points with a 95\% confidence interval of $[+0.88,+3.23]$ points, although the individual $4\times$ and $16\times$ intervals include zero.
Under the joint OLMo--MuSiQue shift, the equal-ratio difference is smaller at $+0.72$ points with a 95\% confidence interval of $[-0.83,+2.34]$ points.
The Reverse anchor therefore provides a useful finite-budget bias across the individual scorer and dataset shifts, while its additional advantage over an extra random candidate becomes weaker when both axes change simultaneously.

\begin{table}[t]
\centering
\caption{
\textbf{Paired retention differences between MORSE-5 and RandomSearch-5.}
Values in brackets denote paired example-level bootstrap 95\% confidence intervals.
}
\label{tab:generalization_rs_ci}
\small
\setlength{\tabcolsep}{3.8pt}
\renewcommand{\arraystretch}{1.08}
\begin{tabular}{@{}llccc@{}}
\toprule
\textbf{Scorer}
& \textbf{Dataset}
& $\mathbf{4\times}$
& $\mathbf{8\times}$
& $\mathbf{16\times}$ \\
\midrule
OLMo
& HotpotQA
& $+.0343\,[.0075,.0628]$
& $+.0432\,[.0158,.0727]$
& $+.0538\,[.0280,.0832]$ \\
Qwen
& MuSiQue
& $+.0123\,[-.0062,.0305]$
& $+.0320\,[.0140,.0508]$
& $+.0178\,[-.0040,.0412]$ \\
OLMo
& MuSiQue
& $-.0103\,[-.0290,.0075]$
& $+.0178\,[-.0095,.0438]$
& $+.0142\,[-.0075,.0378]$ \\
\bottomrule
\end{tabular}
\end{table}

\begin{table}[t]
\centering
\caption{
\textbf{Paired retention differences between MORSE-5 and Reverse.}
Values in brackets denote paired example-level bootstrap 95\% confidence intervals.
}
\label{tab:generalization_reverse_ci}
\small
\setlength{\tabcolsep}{3.8pt}
\renewcommand{\arraystretch}{1.08}
\begin{tabular}{@{}llccc@{}}
\toprule
\textbf{Scorer}
& \textbf{Dataset}
& $\mathbf{4\times}$
& $\mathbf{8\times}$
& $\mathbf{16\times}$ \\
\midrule
OLMo
& HotpotQA
& $+.0917\,[.0452,.1384]$
& $+.0700\,[.0297,.1100]$
& $+.0548\,[.0207,.0910]$ \\
Qwen
& MuSiQue
& $+.0607\,[.0233,.0995]$
& $+.0447\,[.0028,.0892]$
& $+.0712\,[.0370,.1062]$ \\
OLMo
& MuSiQue
& $+.1207\,[.0687,.1740]$
& $+.1123\,[.0602,.1660]$
& $+.1182\,[.0712,.1675]$ \\
\bottomrule
\end{tabular}
\end{table}

\paragraph{Supporting-content retention on MuSiQue.}
Because SP-R counts a supporting paragraph as retained once any sentence survives, we additionally evaluate Supporting-Paragraph Token Retention (SP-TR).
This diagnostic measures how much content originating from gold supporting paragraphs remains after compression.
The results are shown in Table~\ref{tab:musique_sptr}.

\begin{table}[t]
\centering
\caption{
\textbf{Supporting-Paragraph Token Retention on MuSiQue.}
Higher is better.
}
\label{tab:musique_sptr}
\small
\setlength{\tabcolsep}{6.0pt}
\renewcommand{\arraystretch}{1.08}
\begin{tabular}{@{}llccc@{}}
\toprule
\textbf{Scorer}
& \textbf{Method}
& $\mathbf{4\times}$
& $\mathbf{8\times}$
& $\mathbf{16\times}$ \\
\midrule
Qwen
& Reverse
& 0.4218 & 0.3152 & 0.2038 \\
&
RandomSearch-5
& 0.4538 & 0.3128 & 0.2091 \\
&
MORSE-5
& 0.4665 & 0.3365 & 0.2245 \\
\midrule
OLMo
& Reverse
& 0.3061 & 0.2055 & 0.1365 \\
&
RandomSearch-5
& 0.3672 & 0.2327 & 0.1548 \\
&
MORSE-5
& 0.3652 & 0.2415 & 0.1637 \\
\bottomrule
\end{tabular}
\end{table}

The SP-TR results closely follow the SP-R pattern.
Under Qwen, MORSE retains more supporting-paragraph token mass than both Reverse and RandomSearch-5 at all three ratios.
Under OLMo, MORSE substantially exceeds Reverse and exceeds RandomSearch-5 at $8\times$ and $16\times$, while the two search methods are nearly tied at $4\times$.
This agreement indicates that the MuSiQue results are not an artifact of counting a paragraph as preserved after retaining only a minimal amount of its content.

\paragraph{Downstream QA.}
Answer F1 and EM are reported in Table~\ref{tab:generalization_full}.
The downstream effects are less uniform than evidence retention, particularly for MuSiQue.
For example, under OLMo--MuSiQue at $16\times$, MORSE improves SP-R over RandomSearch-5 but decreases Answer F1.
This is consistent with our main experiments, where supporting-evidence retention is a direct measure of preserved annotated evidence but is not a complete surrogate for downstream answer quality.
We therefore use the generalization experiment primarily to assess whether the evidence-preservation behavior of MORSE transfers across compression scorers and datasets.

\paragraph{Summary.}
The generalization results separate two roles within MORSE.
Compression-aware search remains consistently beneficial relative to the static Reverse anchor across every tested scorer--dataset combination.
The Reverse anchor additionally improves finite-budget search relative to compute-matched random exploration under the original condition and under the individual scorer and dataset shifts.
Its marginal contribution becomes smaller under the simultaneous OLMo--MuSiQue shift, where MORSE and RandomSearch-5 obtain statistically indistinguishable retention.
These results support compression-aware ordering as the more broadly transferable component, with Reverse providing a useful but distribution-dependent evidence-first search bias.

\subsection{Context-Count Scaling}
\label{app:context_scaling}

We evaluate how the ordering methods scale as the retrieved context collection grows.
The experiment uses the fixed SAME-100 HotpotQA population and the primary 1P compressor.
For each example, we construct nested context collections with
$n\in\{10,20,30,40,50\}$.
All native contexts and gold supporting evidence are preserved, and additional non-supporting distractors are appended using a single deterministic augmentation sequence, such that the smaller context collections are prefixes of the larger ones.
The $n=20$ condition is identical to the accepted TRUE-20 construction used elsewhere in the paper.

We evaluate Original, Reverse, \textsc{RandomSearch-5}, and \texttt{MORSE-5} under $4\times$, $8\times$, and $16\times$ compression.
\textsc{RandomSearch-5} and \texttt{MORSE-5} use the same five deterministic seeds (42--46), with the first four random permutations shared exactly between their candidate pools.
All other scoring, compression, canonicalization, and downstream evaluation settings are unchanged from the main experiments.
For each context count, the compression budget is computed from the enlarged input using the same nominal compression ratio.
Consequently, increasing the number of contexts also increases the absolute retained-token budget, so this experiment evaluates robustness to larger context collections rather than holding the absolute compression budget fixed.

Table~\ref{tab:context_scaling_sfr} reports the complete evidence-retention results.
\texttt{MORSE-5} outperforms Reverse in all 15 evaluated settings, with paired improvements ranging from 3.85 to 17.15 SF-R points and every 95\% confidence interval strictly above zero.
The comparison with \textsc{RandomSearch-5} is more regime-dependent.
\texttt{MORSE-5} has significantly higher SF-R at $8\times$ with 10 and 20 contexts and at $16\times$ with 20 and 30 contexts.
At $4\times$ with 50 contexts, \textsc{RandomSearch-5} is significantly higher, while the remaining ten comparisons have confidence intervals overlapping zero.
These results indicate that compression-aware global exploration is the more robust scaling component, while the Reverse anchor provides a finite-search-budget advantage whose magnitude depends on the compression regime and context collection.

\begin{table*}[t]
\centering
\caption{
\textbf{Context-count scaling on HotpotQA under 1P compression.}
SF-R is evaluated on the fixed SAME-100 population.
\texttt{MORSE-5} and \textsc{RandomSearch-5} are averaged over seeds 42--46.
The final two columns report paired SF-R differences with 95\% bootstrap confidence intervals over examples.
}
\label{tab:context_scaling_sfr}
\scriptsize
\setlength{\tabcolsep}{3.4pt}
\renewcommand{\arraystretch}{1.08}
\begin{tabular}{@{}ccrrrrll@{}}
\toprule
\textbf{Ratio}
& \textbf{$n$}
& \textbf{Original}
& \textbf{Reverse}
& \textbf{RandomSearch-5}
& \textbf{MORSE-5}
& \textbf{MORSE--RandomSearch}
& \textbf{MORSE--Reverse} \\
\midrule
$4\times$
& 10 & 0.5765 & 0.6780 & 0.7326 & 0.7230
& $-0.0096$ [$-0.0320$, $0.0125$]
& $+0.0450$ [$0.0035$, $0.0867$] \\
& 20 & 0.5793 & 0.6572 & 0.7622 & 0.7780
& $+0.0158$ [$-0.0002$, $0.0337$]
& $+0.1208$ [$0.0733$, $0.1687$] \\
& 30 & 0.5627 & 0.6397 & 0.7735 & 0.7708
& $-0.0026$ [$-0.0242$, $0.0192$]
& $+0.1312$ [$0.0835$, $0.1788$] \\
& 40 & 0.5477 & 0.6238 & 0.8015 & 0.7953
& $-0.0062$ [$-0.0280$, $0.0168$]
& $+0.1715$ [$0.1232$, $0.2225$] \\
& 50 & 0.5560 & 0.6297 & 0.7972 & 0.7693
& $-0.0279$ [$-0.0539$, $-0.0023$]
& $+0.1396$ [$0.0959$, $0.1859$] \\
\midrule
$8\times$
& 10 & 0.4457 & 0.5543 & 0.5825 & 0.6103
& $+0.0278$ [$0.0022$, $0.0551$]
& $+0.0560$ [$0.0203$, $0.0935$] \\
& 20 & 0.4982 & 0.5735 & 0.6547 & 0.6827
& $+0.0280$ [$0.0060$, $0.0527$]
& $+0.1092$ [$0.0634$, $0.1563$] \\
& 30 & 0.4865 & 0.5710 & 0.6889 & 0.6841
& $-0.0048$ [$-0.0260$, $0.0180$]
& $+0.1131$ [$0.0686$, $0.1562$] \\
& 40 & 0.4882 & 0.5843 & 0.7063 & 0.7143
& $+0.0081$ [$-0.0163$, $0.0337$]
& $+0.1300$ [$0.0858$, $0.1757$] \\
& 50 & 0.4940 & 0.5843 & 0.7018 & 0.7078
& $+0.0060$ [$-0.0173$, $0.0309$]
& $+0.1235$ [$0.0798$, $0.1685$] \\
\midrule
$16\times$
& 10 & 0.2803 & 0.3957 & 0.4163 & 0.4342
& $+0.0178$ [$-0.0005$, $0.0368$]
& $+0.0385$ [$0.0047$, $0.0713$] \\
& 20 & 0.3932 & 0.5093 & 0.5202 & 0.5761
& $+0.0559$ [$0.0263$, $0.0892$]
& $+0.0668$ [$0.0352$, $0.0993$] \\
& 30 & 0.4282 & 0.5252 & 0.5836 & 0.6184
& $+0.0347$ [$0.0102$, $0.0619$]
& $+0.0932$ [$0.0538$, $0.1326$] \\
& 40 & 0.4257 & 0.5285 & 0.6081 & 0.6197
& $+0.0116$ [$-0.0112$, $0.0367$]
& $+0.0912$ [$0.0487$, $0.1338$] \\
& 50 & 0.4273 & 0.5252 & 0.6185 & 0.6412
& $+0.0227$ [$-0.0025$, $0.0499$]
& $+0.1161$ [$0.0764$, $0.1556$] \\
\bottomrule
\end{tabular}
\end{table*}

Table~\ref{tab:context_scaling_qa} reports the corresponding downstream Answer F1 and EM results.
The downstream trends are less uniform than SF-R, consistent with the main experiments: improved evidence retention does not guarantee an improvement in generated answers for every individual setting.
Nevertheless, the stronger-compression conditions generally show competitive or improved downstream performance for compression-aware search relative to static ordering.

\begin{table*}[t]
\centering
\caption{
\textbf{Downstream QA for the HotpotQA context-count scaling experiment.}
Each entry reports Answer F1 / EM.
\texttt{MORSE-5} and \textsc{RandomSearch-5} values are five-seed means.
}
\label{tab:context_scaling_qa}
\scriptsize
\setlength{\tabcolsep}{5.0pt}
\renewcommand{\arraystretch}{1.08}
\begin{tabular}{@{}ccllll@{}}
\toprule
\textbf{Ratio}
& \textbf{$n$}
& \textbf{Original}
& \textbf{Reverse}
& \textbf{RandomSearch-5}
& \textbf{MORSE-5} \\
\midrule
$4\times$
& 10 & .5466 / .460 & .6159 / .530 & .6421 / .538 & .6358 / .536 \\
& 20 & .5649 / .470 & .5965 / .500 & .6558 / .550 & .6682 / .572 \\
& 30 & .5559 / .460 & .6040 / .500 & .6478 / .546 & .6437 / .540 \\
& 40 & .5389 / .450 & .6032 / .500 & .6475 / .552 & .6228 / .526 \\
& 50 & .5404 / .440 & .5817 / .490 & .6214 / .522 & .6315 / .536 \\
\midrule
$8\times$
& 10 & .5241 / .430 & .5997 / .500 & .6016 / .498 & .6124 / .506 \\
& 20 & .5020 / .420 & .5900 / .500 & .5833 / .486 & .6082 / .516 \\
& 30 & .5034 / .410 & .6040 / .500 & .6112 / .502 & .6132 / .504 \\
& 40 & .4854 / .400 & .5898 / .490 & .6429 / .544 & .6410 / .536 \\
& 50 & .5029 / .420 & .5945 / .500 & .6098 / .516 & .6064 / .514 \\
\midrule
$16\times$
& 10 & .4290 / .340 & .5194 / .400 & .5096 / .416 & .5172 / .408 \\
& 20 & .5049 / .420 & .5628 / .470 & .5600 / .464 & .5804 / .484 \\
& 30 & .5051 / .420 & .5660 / .450 & .5951 / .502 & .5978 / .500 \\
& 40 & .4894 / .410 & .5370 / .460 & .5898 / .508 & .5948 / .514 \\
& 50 & .4762 / .390 & .5458 / .450 & .6031 / .514 & .6074 / .514 \\
\bottomrule
\end{tabular}
\end{table*}

\end{document}